\pdfoutput=1
\documentclass[11pt]{article}

\usepackage{acl}
 
\usepackage{times}
\usepackage{latexsym}

\usepackage[T1]{fontenc}
\usepackage[utf8]{inputenc}

\usepackage{microtype}

\usepackage{amsmath}
\usepackage{graphicx}
\usepackage{float}
\usepackage{subcaption}
\usepackage{multicol}
\usepackage{booktabs}
\usepackage{multirow}
\usepackage[bottom]{footmisc}
\usepackage{cancel}
\usepackage{caption}
\usepackage{subcaption}
\usepackage{pifont}% http://ctan.org/pkg/pifont
\newcommand{\cmark}{\ding{51}}%
\usepackage{enumerate}

\title{Podcast Summary Assessment: A Resource for Evaluating Summary Assessment Methods}

\author{Potsawee Manakul \textnormal{ and } Mark J. F. Gales \\
  Department of Engineering, University of Cambridge \\
  \texttt{pm574@cam.ac.uk, mjfg@eng.cam.ac.uk}}

\begin{document}
\maketitle
\begin{abstract}
% 1. General Summary Assessment
% 2. New dataset
% 3. Our approach -- benchmark & supervised
% 4. New challenge -- selecting good training examples
Automatic summary assessment is useful for both machine-generated and human-produced summaries. Automatically evaluating the summary text given the document enables, for example, summary generation system development and detection of inappropriate summaries. Summary assessment can be run in a number of modes: ranking summary generation systems; ranking summaries of a particular document; and estimating the quality of a document-summary pair on an absolute scale. Existing datasets with annotation for summary assessment are usually based on news summarization datasets such as CNN/DailyMail or XSum. In this work, we describe a new dataset, the podcast summary assessment corpus, a collection of podcast summaries that were evaluated by human experts at TREC2020. Compared to existing summary assessment data, this dataset has two unique aspects: (i) long-input, speech podcast based, documents; and (ii) an opportunity to detect inappropriate reference summaries in podcast corpus. First, we examine existing assessment methods, including model-free and model-based methods, and provide benchmark results for this long-input summary assessment dataset. Second, with the aim of filtering reference summary-document pairings for training, we apply summary assessment for data selection. The experimental results on these two aspects provide interesting insights on the summary assessment and generation tasks. The podcast summary assessment data is available.\footnote{Data is available at \url{https://github.com/potsawee/podcast_summary_assessment} under the CC-BY-4.0 license.} 

% Podcast descriptions have been used as the ground-truth summaries in training summary generation. However, human evaluation has revealed that only half of the podcast descriptions are graded excellent or good, while the rest of the descriptions are graded fair or bad. This work addresses this issue by removing poor quality descriptions in training summary generation models, and we provide baseline results based on summary assessment methods. 

%  However, the nature of machine-generated and human summaries can be considerably different. In this paper, we provide an overview of existing datasets and approaches for both tasks. We also provide a new dataset for the model-generated summary assessment task.
\end{abstract}

\section{Introduction}
Summarization or summary generation aims to compress a document into a concise summary that conveys the important information, while summary assessment or evaluation aims to provide the quality of the summary text given the document. With the advances in deep learning, a variety of automatic summary generation models have been proposed \cite{see-etal-2017-get, lewis-etal-2020-bart, zhang2020pegasus}. However, there is less attention on automatic summary assessment. 

Firstly, automatic assessment such as ROUGE \cite{lin-2004-rouge} allows researchers to quickly compare and rank summary generation models as it has been shown to have a high/moderate correlation with human judgements at the system-level. Secondly, automatic assessment can also be applied to rank a set of summaries for the document, i.e. summary-level evaluation. The definitions of system-level and summary-level are provided in Section \ref{section:metric_definition}. Thirdly, instead of ranking, another assessment task is to evaluate the quality of a document-summary pair on an absolute scoring scale. This is a regression task which has applications such as assessing summaries of English learners  \cite{xia-etal-2019-automatic}, or selecting good document-summary pairs for training generation systems.

In this work, we compile and release summaries of podcasts and associated human judgements from the Spotify Podcast Challenge at TREC2020 \cite{jones_trec2020}, which is based on podcast data of more than 100,000 episodes for training summary generation systems \cite{clifton-etal-2020-100000}. The Podcast Summary Assessment dataset consists of long documents, e.g. the average number of words is more than 6000, meaning that some assessment methods may fail to correlate well with human judgements. Using this new dataset for assessment, we provide benchmark results of standard and recent assessment methods as measured by system-level and summary-level correlations.

In addition, we link the summary assessment task to the summary generation task. Creator-provided podcast descriptions have been used as the reference summaries in training summary generation models \cite{manakul2020cued_speech}; however, human evaluation suggests that up to half of the descriptions are judged as only fair or bad (see Section \ref{section:data_human_annotation}). Thus, it is a challenge to select appropriate and high-quality training examples for the generation task. In this work, we propose using summary assessment to tackle this data selection problem, and we provide baseline results and insights based on supervised assessment models. The main contributions of this paper are:

\begin{itemize}
    \item We assemble and release \textit{Podcast Summary Assessment} -- a summary assessment dataset based on a large podcast summarization data from the podcast challenge at TREC2020. The data provides a diverse assessment resource beyond the scope of news articles.
    
    \item We provide benchmark results including several assessment methods on the new dataset.
    \item We link the assessment task to the generation task, and we provide baseline results.
\end{itemize}

\section{Related Work: Assessment Methods}
\label{section:assessment_methods}
Our notation is $\mathbf{x}$ = document, $\mathbf{y}$ = candidate summary, $\mathbf{y}^*$ = reference summary, and $z$ = quality of the summary. We categorize summary assessment methods by: first, $f(\mathbf{y},\mathbf{y}^*)$ v.s. $f(\mathbf{y},\mathbf{x})$ i.e. whether the summary is compared against the document or the reference summary; second, unsupervised approach v.s. supervised approach. In this section, we provide the details of methods used in this work. A literature review of recent summary assessment or evaluation methods can be found in \citet{koto2022ffci}.
\subsection{Summary and Reference $f(\mathbf{y},\mathbf{y}^*)$}
\label{section:f_xy_ref}
Typically, datasets for developing summary generation systems contain a set of documents $\mathbf{X}=\{\mathbf{x}^{(1)},\mathbf{x}^{(2)},...\}$ and reference summaries $\mathbf{Y}^*=\{\mathbf{y}^{*(1)},\mathbf{y}^{*(2)},...\}$. Generation systems are trained to maximise the likelihood of the reference summaries such that $\theta_{\tt{ml}} = \text{argmax}_{\theta} [ P(\mathbf{Y}^*|\mathbf{X};\theta) ]$. Consequently, standard summary assessment methods take the form $f(\mathbf{y},\mathbf{y}^*)$.

\subsubsection{Unsupervised $f(\mathbf{y},\mathbf{y}^*)$}
By far the most commonly used $f(\mathbf{y},\mathbf{y}^*)$ method is ROUGE \cite{lin-2004-rouge}, which is model-free and based on the n-gram overlap between $\mathbf{y}$ and $\mathbf{y}^*$. Other variants of n-gram based methods include BLEU \cite{papineni-etal-2002-bleu}. Despite its robustness, n-gram based methods cannot take in account word semantic. Model-based word-level representation matching such as BERTScore \cite{BERTScore} or MoverScore \cite{zhao-etal-2019-moverscore} are proposed to incorporate word semantic. This idea could be extended into sentence-level representation matching such as Sentence-BERT \cite{reimers-gurevych-2019-sentence}. Rather than n-gram matching or representation matching, triple matching has also been proposed \cite{goodrich_triple}.

\subsubsection{Supervised $f(\mathbf{y},\mathbf{y}^*)$}
Methods such as BLEURT \cite{sellam-etal-2020-bleurt} or COMET \cite{rei-etal-2020-comet} are trained to predict human scores given $\mathbf{y}$ and $\mathbf{y}^*$. However, it is tedious to collect both human scores and $\mathbf{y}^*$, making it less practical. Thus, we omit this type of approach.

\subsection{Summary and Document $f(\mathbf{y},\mathbf{x})$}
\label{section:f_xy_doc}
In a practical scenario as such assessing human's summarization skill without reference summary $\mathbf{y}^*$ being available, the document $\mathbf{x}$ has to be used in assessing the summary $\mathbf{y}$.

% Because $\mathbf{x}$ contains richer information than $\mathbf{y}^*$, when comparing summary $\mathbf{y}$ against $\mathbf{x}$, existing approaches generally focus on assessing particular aspects of summary quality rather than holistically. 

\subsubsection{Unsupervised $f(\mathbf{y},\mathbf{x})$}
\noindent \textbf{Question-Answering.}
To assess the faithfulness aspect, \citet{wang-etal-2020-asking} proposed QAGS, the first QA-based method. Given $\mathbf{y}$, noun-phrases are extracted. For each noun-phrase, generate a question through $\text{noun}+\mathbf{y} \rightarrow \text{question}$, and the answer conditioned on $\mathbf{x}$ is compared to the answer conditioned on $\mathbf{y}$, e.g. word overlap F1.
\begin{equation}
   \text{QA-score} = \mathop{E}_{Q \sim P(Q|\mathbf{y})} [ D (P(A|Q,\mathbf{x}) , P(A|Q,\mathbf{y})  ] 
   \label{eq:qag}
\end{equation} 
A concurrent and similar QA-based method called FEQA was also proposed by \citet{durmus-etal-2020-feqa}. Because QAGS is a precision-based metric (e.g. it generates questions from $\mathbf{y}$ and checks for consistency against $\mathbf{x}$), \citet{scialom-etal-2021-questeval} proposed QuestEval, which is a combination of QAG-Precision and QAG-Recall. 

% Furthermore, QA-based approaches are also used in improving the faithfulness of summary generation systems \cite{dong-etal-2020-multi, nan-etal-2021-improving} 

\vspace{0.5em}
\noindent \textbf{Entailment.}
Textual entailment task is that given a premise/context $\mathbf{x}$ and hypothesis $\mathbf{y}$, predict one of three possible relations: entail, neutral, contradict. A common training data is Multi-Genre Natural Language Inference (MNLI), which is a crowd-sourced collection of 433k sentence pairs. \citet{maynez-etal-2020-faithfulness} showed that BERT fine-tuned to MNLI achieves the highest Spearman correlation with human judgements on faithfulness and factuality.

\vspace{0.5em}
\noindent Other unsupervised $f(\mathbf{y},\mathbf{x})$ approaches include Language Model Score. For example, \citet{yuan2021bartscore} proposed a conditional LM score
$
    \text{BARTScore} = \sum_{t=1}^M \omega_t \log P(\mathbf{y}_t | \mathbf{y}_{<t}, \mathbf{x}; \theta) 
$
where $\omega_t$ is a weight such as TF-IDF for each word. 

% They also investigate augmenting $\mathbf{x}$ and $\mathbf{y}$ via prompting and fine-tuning $\theta$ to the target task. \\

\subsubsection{Supervised $f(\mathbf{y},\mathbf{x})$}

Supervised approaches require ground-truth scores (human judgements) $\mathbf{Z}^* = \{z^{*(1)},{z}^{*(2)},...\}$  to train regression models $\theta_{\tt reg}$:
\begin{equation}
    \theta_{\tt reg} = \text{argmax}_{\theta} [P(\mathbf{Z}^* | \mathbf{X}, \mathbf{Y}; \theta)]
\end{equation}
For example, \citet{xia-etal-2019-automatic} collected English learners' summaries from a real examination, and have the summaries graded by professional examiners. Kernel Ridge Regression, LSTM, and CNN models were trained using this data.  \citet{bao2020end} trained fully connected, CNN, LSTM, and BERT-based models on \textit{simulated} CNN/DailyMail, BillSum, arXiv, BigPatent data. They created simulated by negative sampling, e.g. random shuffling summaries or word-level summary corruption. Similarly, \citet{kryscinski-etal-2020-evaluating} proposed FactCC metric by fine-tuning BERT classifier on adversarial data to distinguish between faithful and unfaithful summaries. \citet{wu-etal-2020-unsupervised} constructed negative samples with respect to linguistic qualities and informativeness, and they trained BERT-based models using contrasting learning.

% \citet{peyrard-etal-2017-learning} trained a SVR model using the reference summaries $\mathbf{y}^*$ in addition to $\mathbf{x}$ and $\mathbf{y}$ to predict scores on TAC2008/TAC2009 datasets.

\section{Related Work: Data}

DUC 2001-2003\footnote{\url{https://duc.nist.gov/data.html}} and TAC 2008-2010 datasets \cite{dang2008OverviewOT, dang2009OverviewOT} consist of summaries and human evaluation from news articles. Despite the size, the systems in these corpora are extractive and no longer matched current abstractive summarization systems. 

Recently, the summaries of CNN/DailyMail \cite{hermann2015teaching} and XSum \cite{narayan-etal-2018-dont} are annotated to address the lack of summary assessment resource. For example, \citet{maynez-etal-2020-faithfulness} collected annotation for XSum summaries on faithfulness and factuality aspects. QAGS \cite{wang-etal-2020-asking} released annotation for CNNDM and XSum on faithfulness. NeR18 \cite{grusky-etal-2018-newsroom} has human annotation for some of its summaries. RealSum \cite{bhandari-etal-2020-evaluating} and SummEval \cite{fabbri2021summeval} annotated recent advanced summarization systems for CNNDM. In addition, a corpus for summary assessment for English learners was collected \cite{xia-etal-2019-automatic}. Human evaluation assesses the quality of summaries on one or more aspects as follows: 

\begin{itemize}
\item {Informative} (relevance) = how much salient information is presented in the summary, and it should contain little or no redundancy.

\item {Faithfulness} = whether the information in the summary can be inferred by the document. An unfaithful summary contains hallucination, which can be categorized into (i) \textit{intrinsic} hallucination when information is manipulated inaccurately; (ii) \textit{extrinsic} hallucination, which is when information is added.

\item Factuality (consistency) = whether the information in the summary (regardless of its presence in the document) is right or wrong.

\item Fluency = how good the language usage, e.g. no grammatical errors.

\item {Coherence} = collective quality of all sentence, e.g. how well are sentences connected.
\end{itemize}
Overall quality is typically assessed as a combination of the aspects. We summarize existing datasets and their annotation aspects in Table \ref{tab:data}.

\begin{table*}[!t]
\tabcolsep=1mm
  \centering
  \scalebox{0.9}{
  \begin{tabular}{llll}
    \toprule
    Corpus   &Data  &Size$^\dagger$   &Annotation     \\
    \midrule
    TAC2008 &News &2736 (57$\times$48) &Fluency, Relevance, Overall \\
    TAC2009 &News &2420 (55$\times$44) &Fluency, Relevance, Overall \\
    TAC2010 &News &1978 (43$\times$46) &Fluency, Relevance, Overall \\

    XSum Faithfulness  &News (XSum) &2500 (5$\times$500) &Faithfulness, Factuality \\ %\cite{maynez-etal-2020-faithfulness}
    QAGS  &News (CNNDM,XSum) &235, 239  &Faithfulness \\ % \cite{wang-etal-2020-asking}
    % \midrule
    NeR18  &News  &420 (7$\times$60) &Coherence, Fluency, Relevance, Informative \\ % \cite{grusky-etal-2018-newsroom}
    
    RealSum  &News (CNNDM)  &2500 (25$\times$100) &Coverage \\ %\cite{bhandari-etal-2020-evaluating}
    SummEval  &News (CNNDM)  &1600 (16$\times$100) &Coherence, Faithfulness, Fluency, Relevance  \\ %\cite{fabbri2021summeval}
    English Learner  &English Exam &411 &Informative, Coherence, Fluency \\ %\cite{xia-etal-2019-automatic}
    \midrule
    Podcast Summary   &\multirow{2}{*}{Podcast} &\multirow{2}{*}{3580 (20$\times$179)} &4-point scale Overall (Informative \& Fluency)  \\
    Assessment        &      &     &and 8 binary attributes (e.g. names, topic, etc.) \\ 
    \bottomrule
      \end{tabular}}
  \caption{Summary of Datasets. $^\dagger$\#systems$\times$\#documents}
  \label{tab:data}
\end{table*}

\section{Podcast Summary Assessment Data}
The corpus is a collection of podcast summaries generated by recent summarization systems at the Spotify Podcast Challenge at TREC2020 \cite{jones_trec2020}. The summary assessment corpus consists of 179 podcast episodes (i.e. documents). All episodes have summaries from 20 systems (19 summarization systems + 1 creator desccription), and human evaluation was performed by NIST\footnote{\url{https://www.nist.gov/}} assessors for the TREC2020 challenge, resulting in 3580 annotated document-summary pairs in total.

\subsection{Summarization Systems}
20 summarization systems \cite{jones_trec2020, zheng2020two, manakul2020cued_speech, song2020automatic, glasgow_trec, hk_uu_trec} are:

\vspace{0.5em}
\noindent \textbf{Reference}\footnote{Creator-provided description has been used as the reference summary in training podcast summarization systems.}  = R1.
% However, less than half of the creator descriptions are graded with the highest score (see Fig. \ref{fig:grade_creator}).

\vspace{0.5em}
\noindent \textbf{Extractive systems} = E1, E2, E3.

\vspace{0.5em}
\noindent \textbf{Abstractive systems} = A1, A2, A3,..., A16.

\vspace{0.5em}
Extractive systems are based on TextRank \cite{mihalcea-tarau-2004-textrank}, while abstractive systems use a form of deep learning and pre-trained seq2seq models including BART \cite{lewis-etal-2020-bart} and T5 \cite{raffel2020exploring}. Full details of all the systems can be found in \citet{jones_trec2020}.
\begin{table}[!ht]
  \centering
  \begin{tabular}{r|cc}
    \toprule
               &\#sentences    &\#words \\
    \midrule
    Transcript &303{\small$\pm$258}   &6375{\small$\pm$5092} \\
    Summary    &5.9{\small$\pm$9.2}  &98{\small$\pm$75} \\
    \bottomrule
  \end{tabular}
  \caption{Length (Avg.$\pm$Std.) based on \texttt{nltk} tokenizer.}
  \label{tab:length_stat}
\end{table}

\subsection{Human Annotation}
\label{section:data_human_annotation}
The summaries were judged by NIST assessors on a 4-point scale (Excellent-Good-Fair-Bad). An excellent summary should be informative and has no redundancy, and it should be fluent. A bad summary does not convey any salient information (not informative), or not factually correct. More descriptions about the annotation guideline can be found in \citet{jones_trec2020}.

\begin{figure}[!ht]
    \begin{subfigure}[b]{0.49\linewidth}
    \centering
      \includegraphics[width=\linewidth,keepaspectratio]{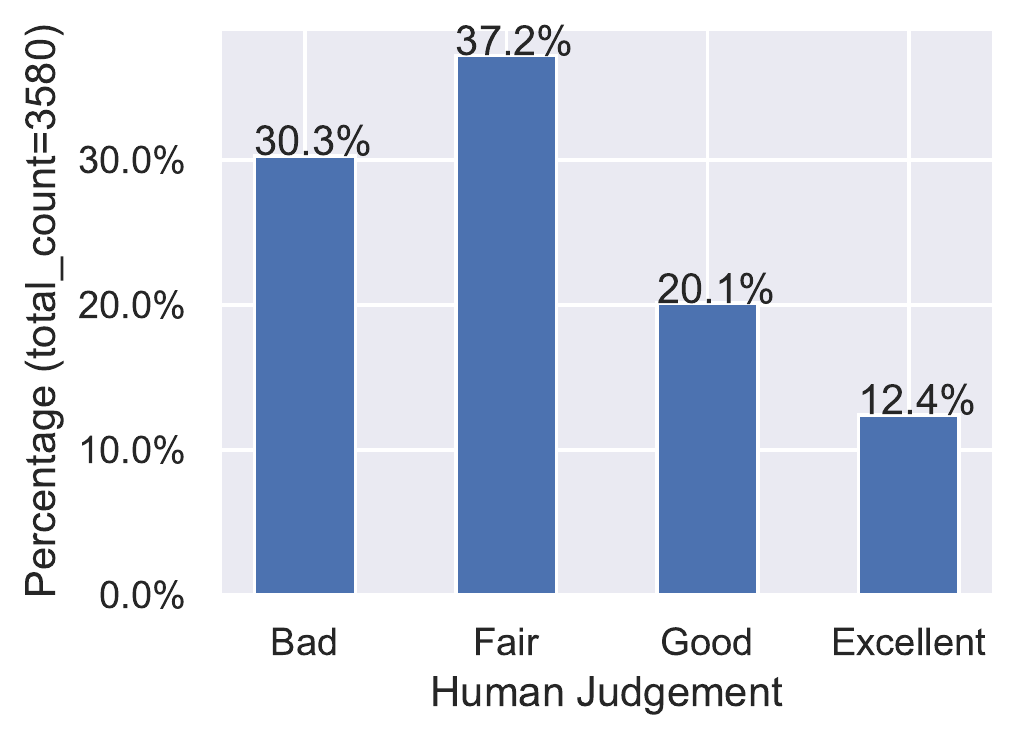}
    \caption{All systems}
    \label{fig:grade_all}
    \end{subfigure}
        % \hfill
    \begin{subfigure}[b]{0.49\linewidth}
    \centering
      \includegraphics[width=\linewidth,keepaspectratio]{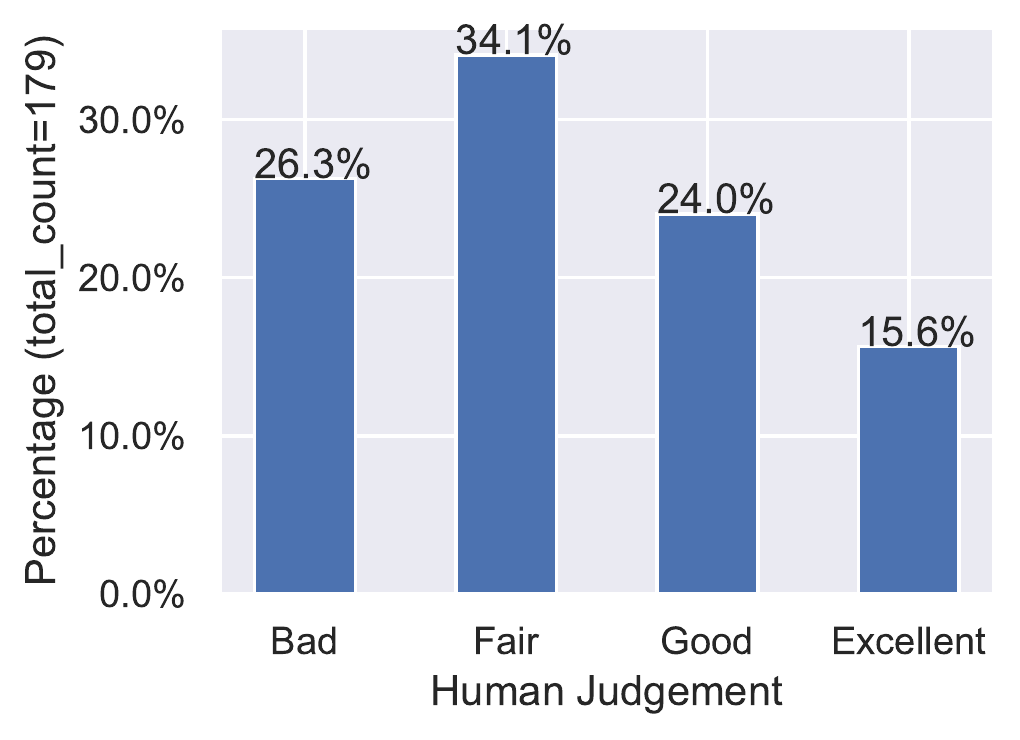}
    \caption{Creator Description}
    \label{fig:grade_creator}
    \end{subfigure}
    \caption{The distribution of human scores.}
    \label{fig:score_distribution}    
\end{figure}

Shown in Fig. \ref{fig:score_distribution} is the distribution of human scores. It can be seen that around a quarter of creator descriptions are graded \textit{Bad}. This result means noisy data in training summarization systems, and it motivates our work in Section \ref{section:sumamry_generation}.

Additionally, the annotation includes 8 binary attributes such as whether the summary contains topic information. This work has not used utilized this annotation. More information can be found in Appendix \ref{appendix:8questions} and \citet{jones_trec2020}.

\section{Assessment Method Evaluation}

\subsection{Evaluation Metrics}
\label{section:metric_definition}
Following the notation in \citet{deutsch-etal-2021-towards}, let $x_i^j$ and $y_i^j$ be two scores of metrics $X$ and $Y$ for the summary output by system $i \in \{1,...,N\}$ on the document $j \in \{1,...,M\}$. Correlations are:
\begin{itemize}
    \item System-level (\textit{aka} Corpus-level)
    \begin{equation}
        \rho = \text{Corr} \left( \left\{ \frac{\sum_j x_i^j}{M} , \frac{\sum_j y_i^j}{M}\right\}_{i=1}^N \right)
    \end{equation}
    \item Summary-level (\textit{aka} Sentence-level)
    \begin{equation}
        \rho = \frac{1}{M} \sum_j \text{Corr} \left( \left\{x_i^j, y_i^j\right\}_{i=1}^N\right)
    \end{equation}
    \item All test examples
    \begin{equation}
        \rho = \text{Corr} \left( \left\{x_i^j, y_i^j\right\}_{i=1, j=1}^{i=N,j=M}  \right)
        \label{eq:correlation_all_pairs}
    \end{equation}    
\end{itemize}
Correlation in Eq. \ref{eq:correlation_all_pairs} is used in Section \ref{section:assessment_for_selection} where all document-summary pairs are evaluated together on an absolute scale. Note that Eq. \ref{eq:correlation_all_pairs} is only applicable when the assessment method gives a score on an absolute scale. For example, the ROUGE score per one document is \textit{not} comparable across different documents, i.e. it is not on an absolute scale.

% For example, say $\text{doc}_1$ and $\text{sum}_1^1,\text{sum}_1^2,\text{sum}_1^3,...$ may be around 0.2, while $\text{doc}_2$ and $\text{sum}_2^1,\text{sum}_2^2,\text{sum}_2^3,...$ may be around 0.9. In words, for document1, the best summary may have a score of 0.2 (e.g. measured by ROUGE), while the worst summary of document2 may have a score of 0.9.

\subsection{Assessment Method Setup}
\label{section:assessment_method_setup}
Implementation details are given in Appendix \ref{section:appendix_implement}.

\vspace{0.2em}
\noindent \textbf{ROUGE} and \textbf{TripleMatching}: ROUGE-1,2,L typically show the same ranking trend, so as a simple unsupervised baseline we report ROUGE-L F1 similar to \citet{jones_trec2020}. Instead of n-gram matching such as ROUGE or BLEU, we follow \citet{goodrich_triple} in extracting a set of triples (Subj-Relation-Obj) from two texts, and we compute the F1-score of the triple overlap.

\vspace{0.5em}
\noindent \textbf{Question-Answering (QA)}: We follow QAG in Eq. \ref{eq:qag} \cite{wang-etal-2020-asking}. For question generation, BART fine-tuned to NewsQA \cite{trischler-etal-2017-newsqa} is used. For question answering, BERT (max \#words = 512) and Longformer (max \#words = 4096) fine-tuned to SQuAD2.0 are used.

\vspace{0.5em}
\noindent \textbf{Entailment}: We train BERT/Longformer on the MNLI corpus \cite{williams-etal-2018-broad}. At inference time,  document $\mathbf{x}$ (context) and summary $\mathbf{y}$ (hypothesis) are concatenated as the input, and the entailment probability is used as the summary score.

\vspace{0.5em}
\noindent \textbf{CNN model}: Due to long documents, we use the sentence-level similarity grid as the input to our CNN model. Document and summary are split into sentences, and each sentence is encoded to a sentence representation via Sentence-BERT \cite{reimers-gurevych-2019-sentence}. Cell ($i,j$) in the similarity grid is cosine similarity between doc-sent$_{i}$ and summary-sent$_{j}$. CNN uses ResNet18 backbone.

\vspace{0.5em}
\noindent \textbf{BERT} \cite{devlin-etal-2019-bert} and \textbf{Longformer} \cite{beltagy2020longformer}: We fine-tune sequence classification weights where the input is $\mathbf{x}$ concatenated by $\mathbf{y}$ and the target is $z$. When $[\mathbf{x};\mathbf{y}]$ exceeds model's max length, we first truncate $\mathbf{x}$. 

In the weakly supervised setting, $z$ is ROUGE-L($\mathbf{y}$,$\mathbf{y}^*$). In the supervised setting, $z$ is human score: \textbf{Excellent=3, Good=2, Fair=1, Bad=0}. Because 3,580 assessment examples is small for training a deep learning model, we perform a 5-fold cross-validation in our supervised training experiments. Also, we perform 5-fold cross-validation 5 times with different data shuffles, and we report the mean of 5 runs (and the standard deviation in Section \ref{section:assessment_for_selection} where we focus on supervised models).

\subsection{Correlation against Human Judgements}

% Commonly used metric for assessing summarization has been ROUGE for more than a decade. It has been shown that despite ROUGE's moderate correlation with human judgement at the \textit{system} level, the correlation at the sentence level is low. In most real use cases, to prevent the failure of automatic summarizers and to assess human's summarization skills (such as English learners), we need a metric that correlates well with human judgement at the \textit{sentence} level. Thus, using our new dataset of summaries generated by modern summarizers, we aim to establish the baseline performance of existing approaches at both corpus and sentence levels. These exisiting approaches are discussed in Section \ref{section:f_xy_ref} and Section \ref{section:f_xy_doc}.

% It is notable that Spearman correlation:
% \begin{itemize}
%     \item QA [B-512] and QA [L-4096] is 0.945.
% \end{itemize}

\begin{figure*}[!t]
    \begin{subfigure}[b]{0.21\linewidth}
    \centering
      \includegraphics[width=\linewidth,keepaspectratio]{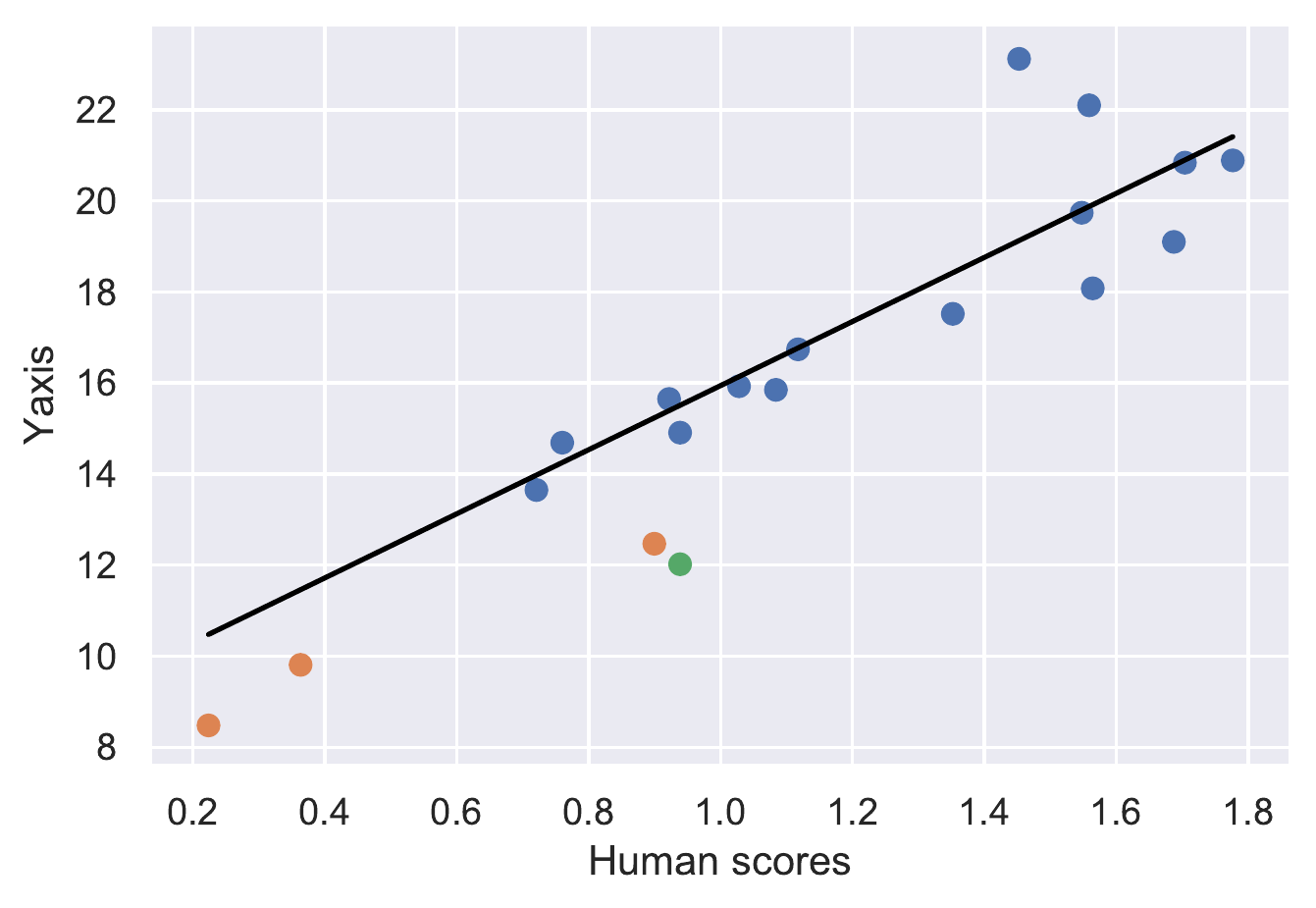}
    \caption{ROUGE-L ($\mathbf{y}$,$\mathbf{y}^*$)}
    \label{fig:rougeL_ref} 
    \end{subfigure}
    \hfill
  \begin{subfigure}[b]{0.21\linewidth}
    \centering
      \includegraphics[width=\linewidth,keepaspectratio]{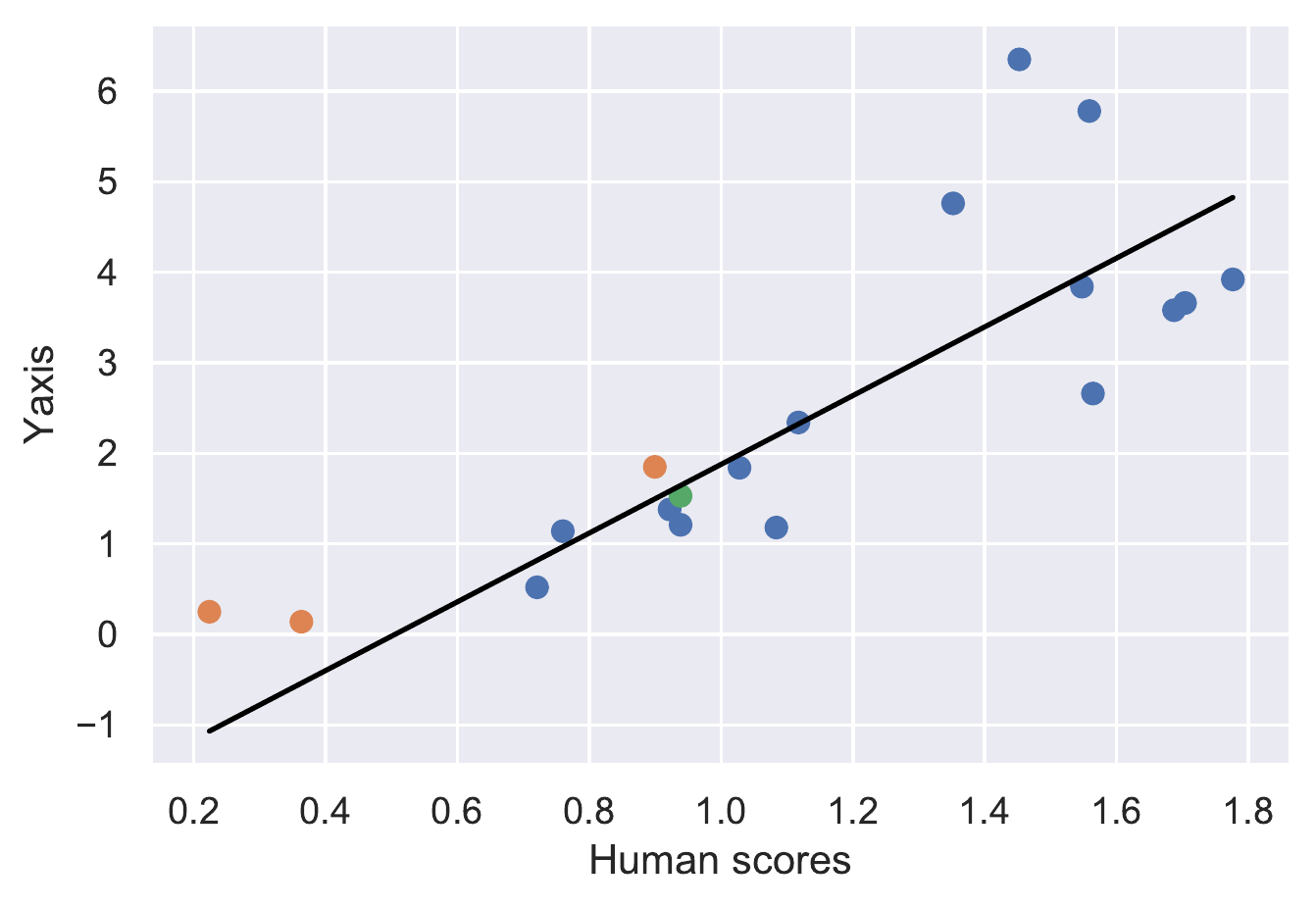}
    \caption{TripleMatch ($\mathbf{y}$,$\mathbf{y}^*$)}
    \label{fig:triple_ref} 
    \end{subfigure}
    \hfill
    \begin{subfigure}[b]{0.21\linewidth}
    \centering
      \includegraphics[width=\linewidth,keepaspectratio]{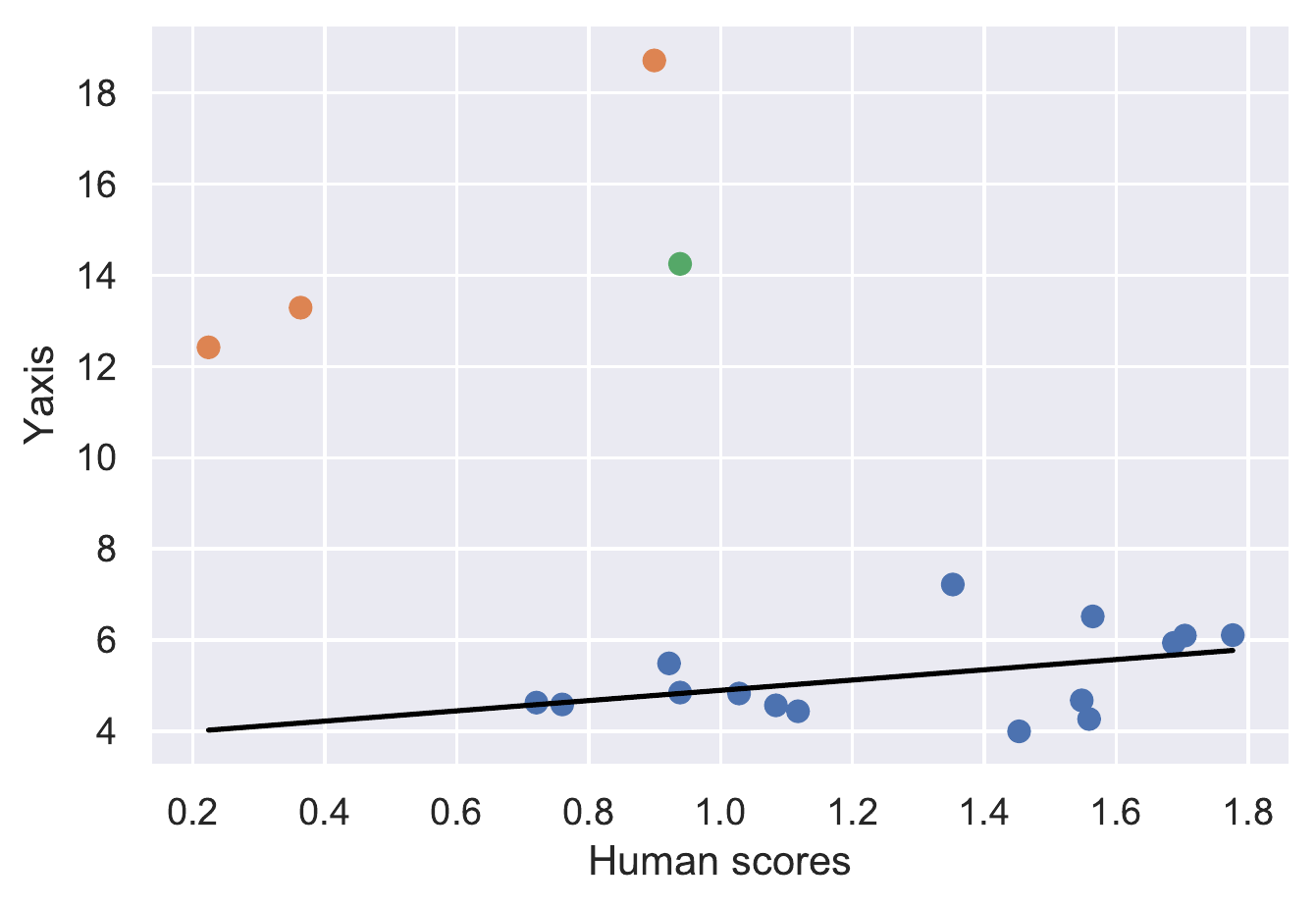}
    \caption{ROUGE-L ($\mathbf{y}$,$\mathbf{x}$)}
    \label{fig:rougeL_doc} 
    \end{subfigure}    
        \hfill
  \begin{subfigure}[b]{0.21\linewidth}
    \centering
      \includegraphics[width=\linewidth,keepaspectratio]{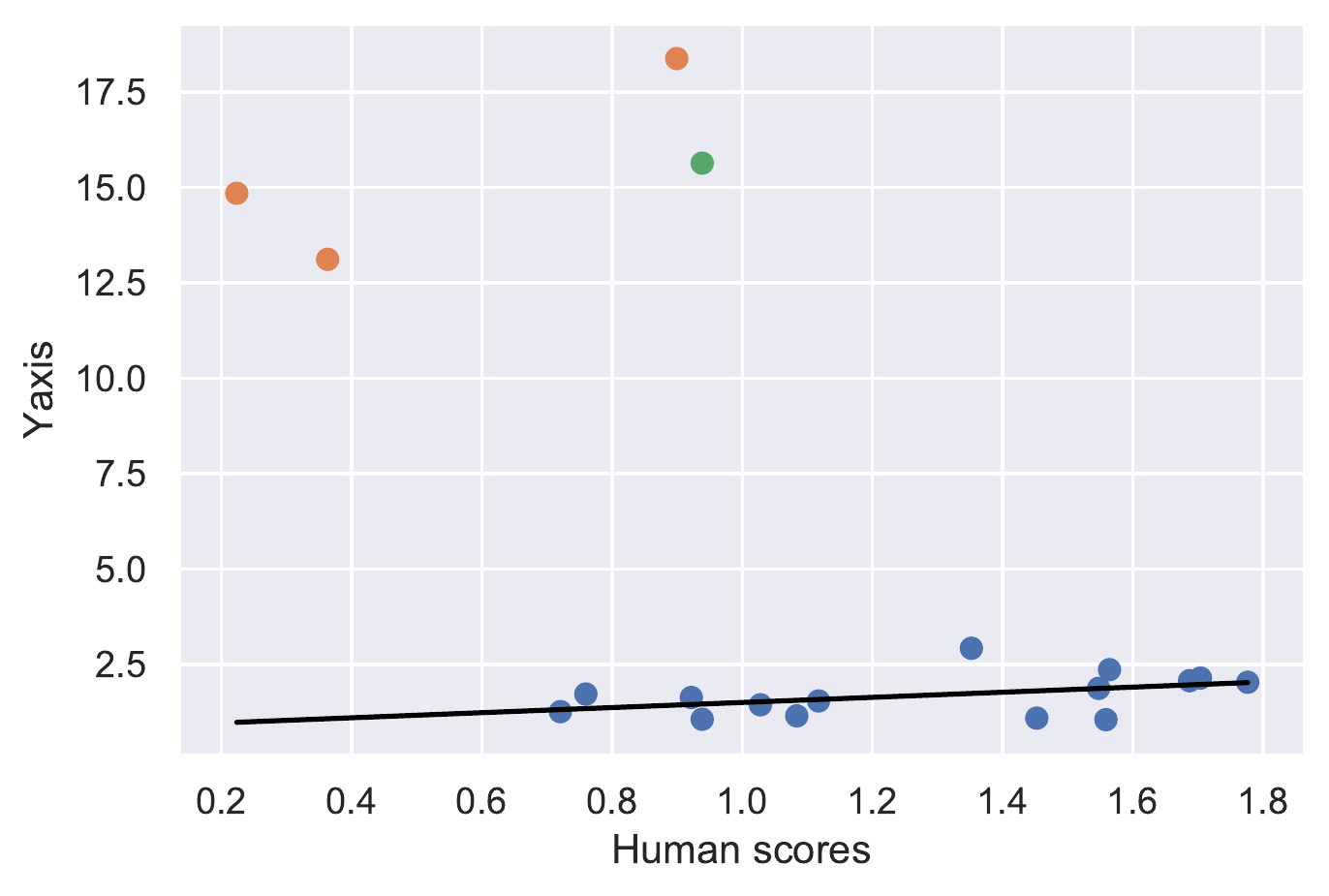}
    \caption{TripleMatch ($\mathbf{y}$,$\mathbf{x}$)}
    \end{subfigure}
  
  \begin{subfigure}[b]{0.21\linewidth}
    \centering
      \includegraphics[width=\linewidth,keepaspectratio]{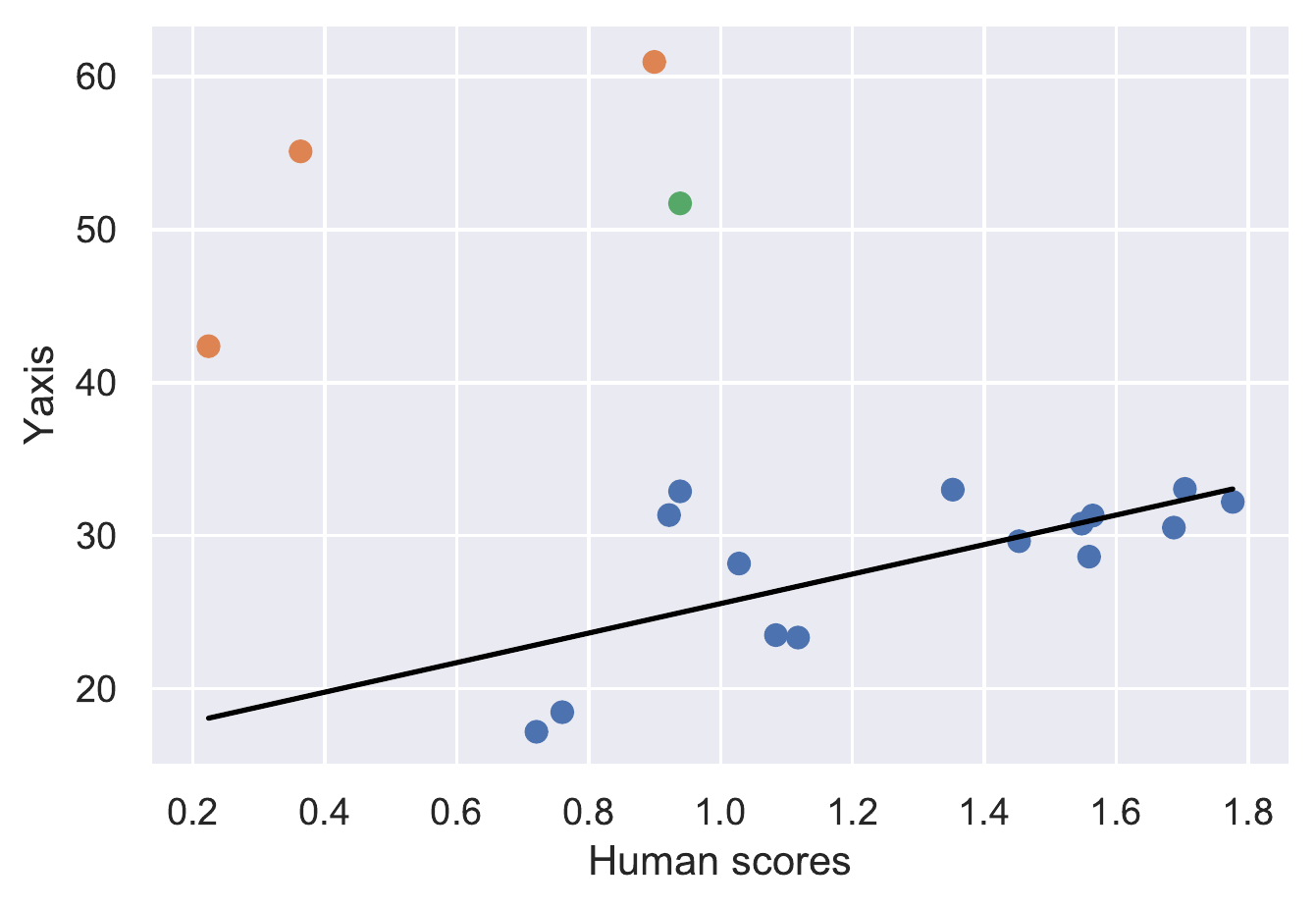}
    \caption{QA [BERT-512]}
    \end{subfigure} 
    \hfill
  \begin{subfigure}[b]{0.21\linewidth}
    \centering
      \includegraphics[width=\linewidth,keepaspectratio]{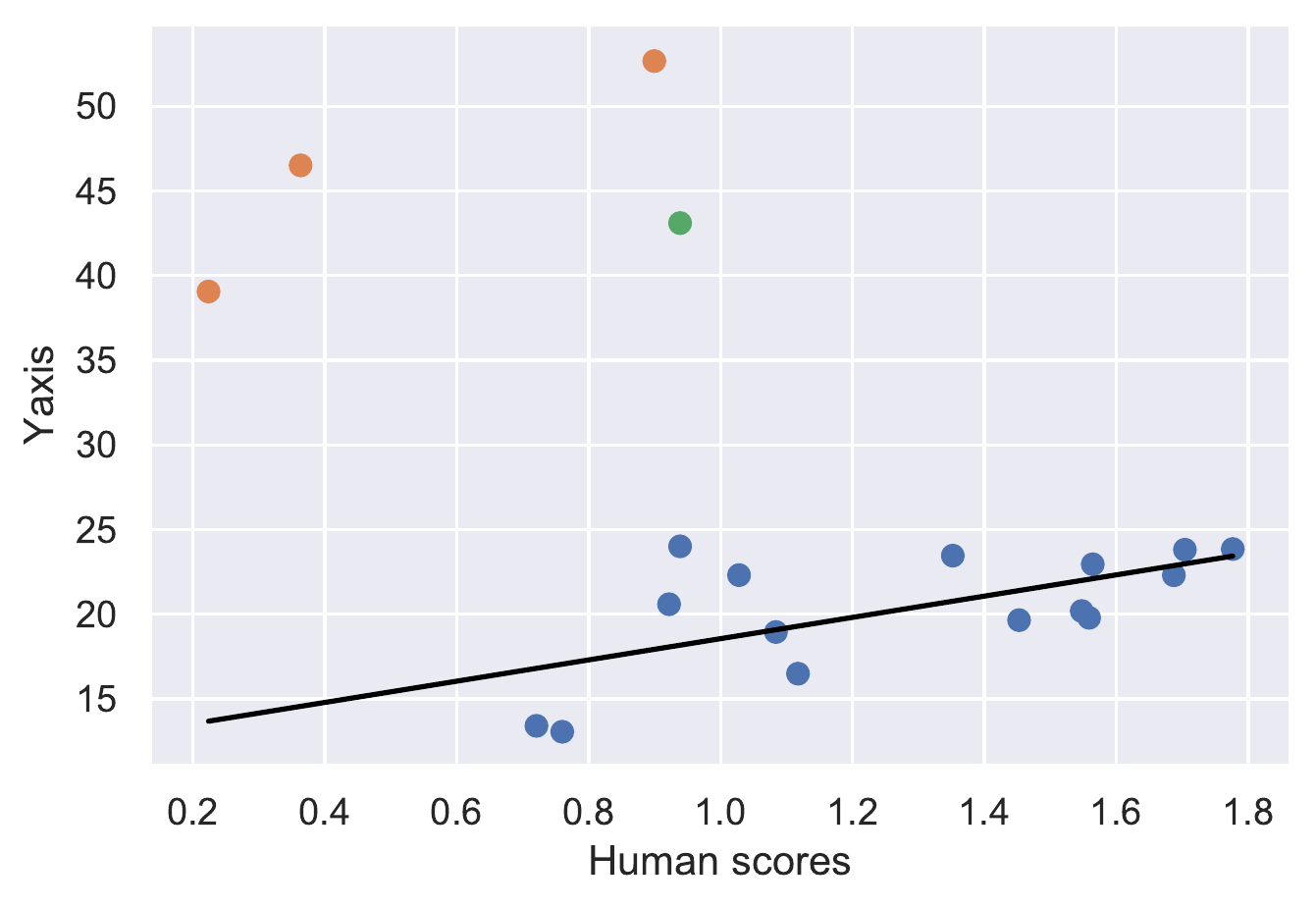}
    \caption{QA [Long-4096]}
    \end{subfigure} 
    \hfill
  \begin{subfigure}[b]{0.21\linewidth}
    \centering
      \includegraphics[width=\linewidth,keepaspectratio]{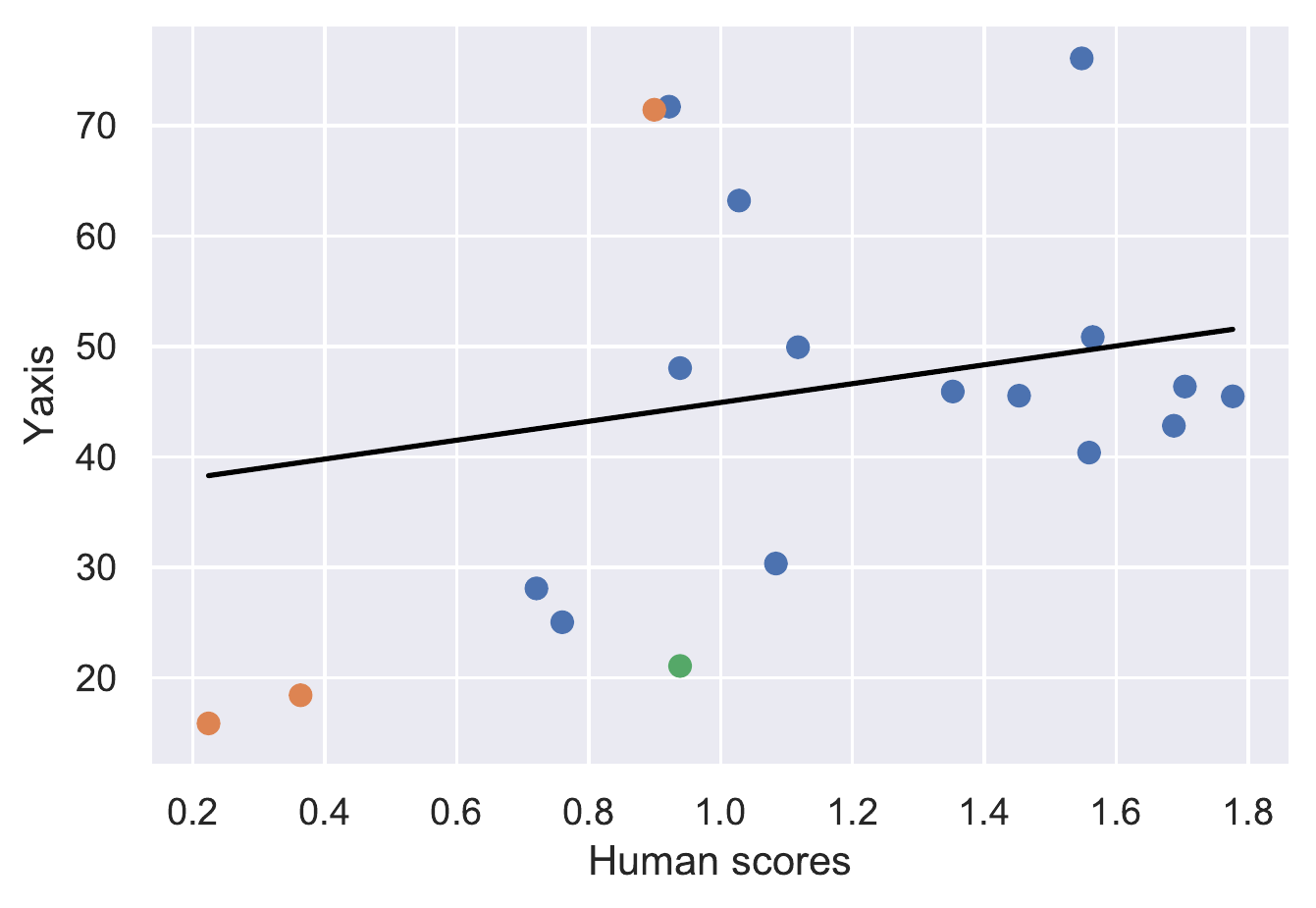}
    \caption{Entail [BERT-512]}
    \end{subfigure} 
    \hfill
  \begin{subfigure}[b]{0.21\linewidth}
    \centering
      \includegraphics[width=\linewidth,keepaspectratio]{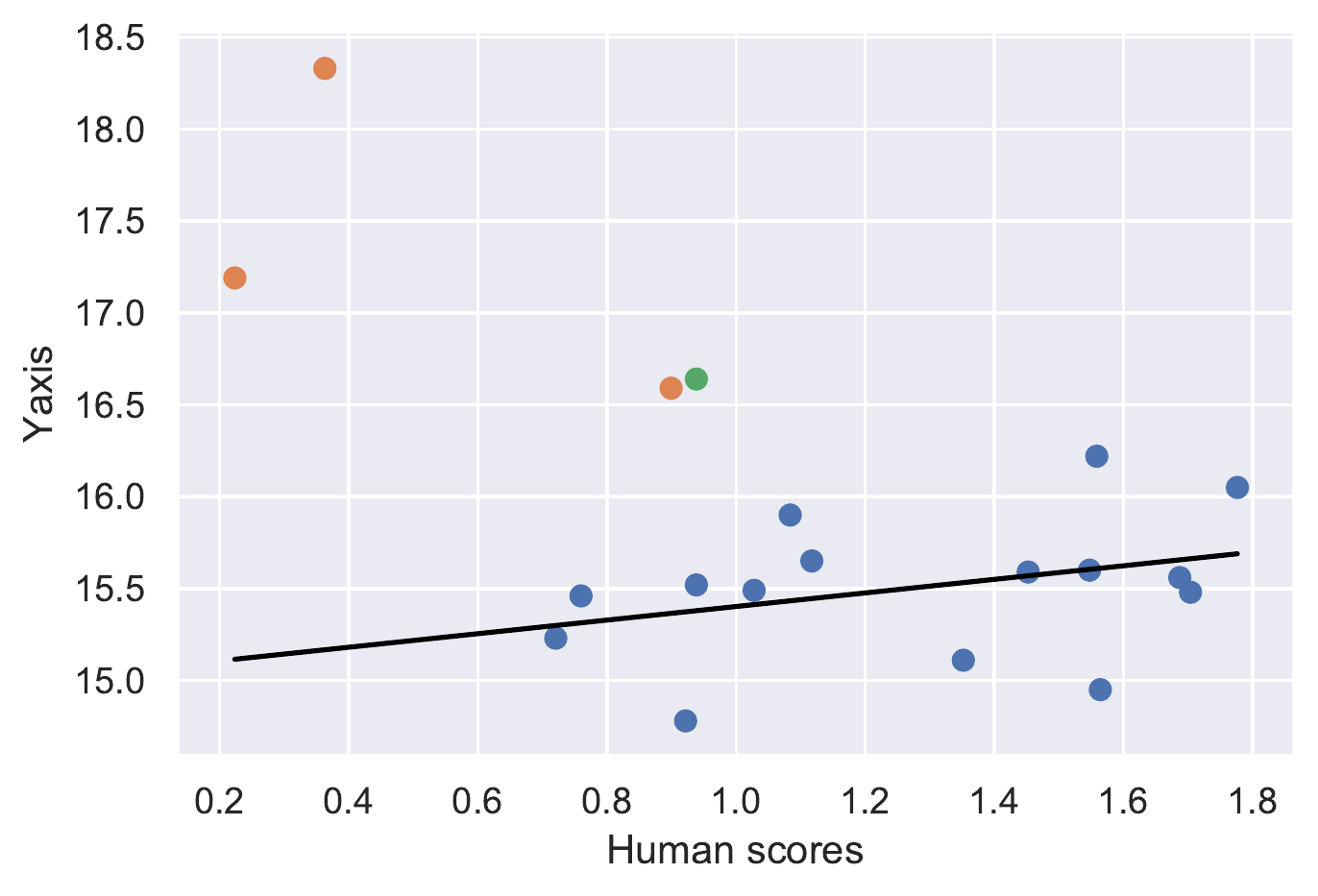}
    \caption{Entail [Long-4096]}
     \label{fig:entail_4096}
    \end{subfigure}
     \hfill
  \begin{subfigure}[b]{0.21\linewidth}
    \centering
      \includegraphics[width=\linewidth,keepaspectratio]{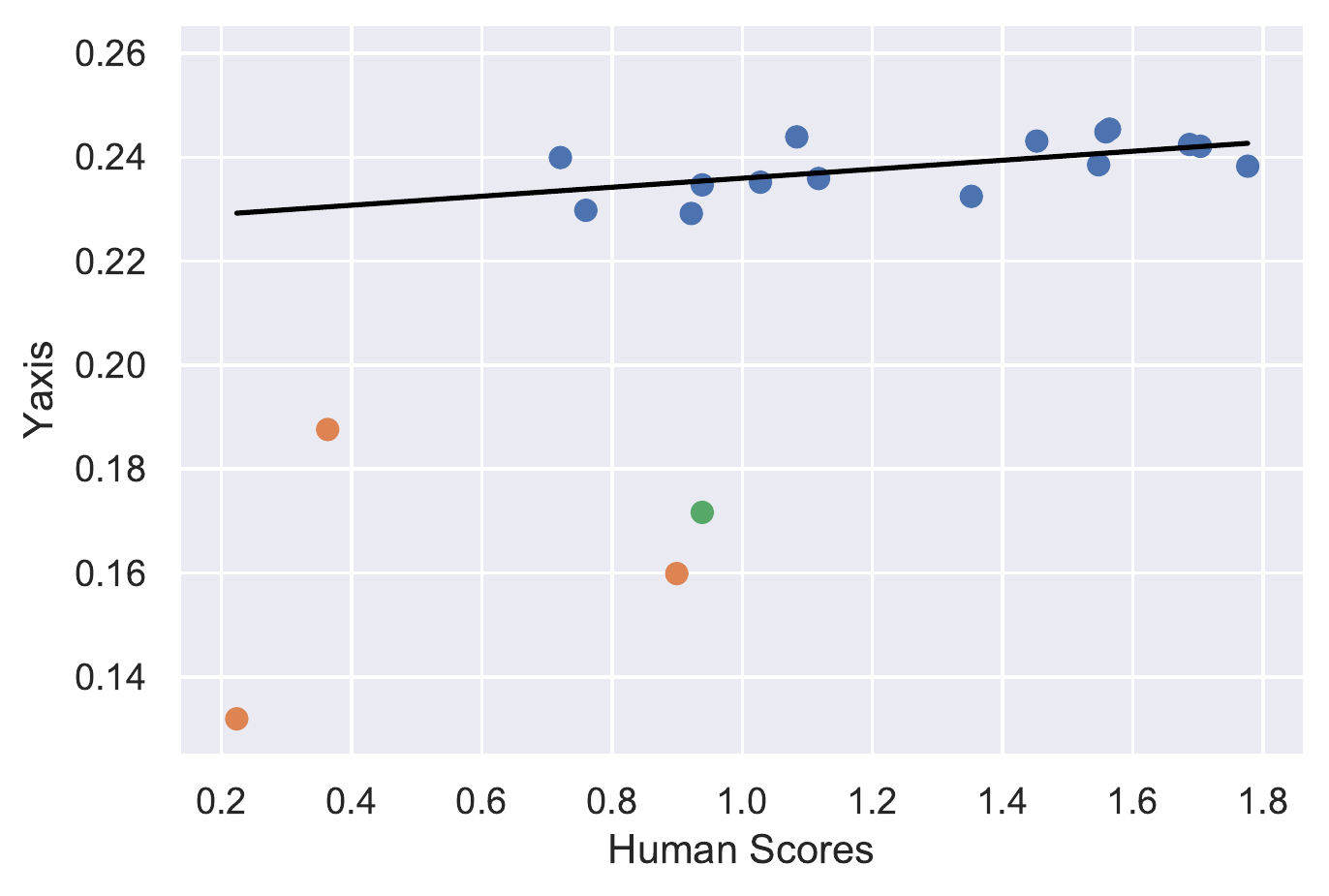}
    \caption{CNN (ROUGE)}
    \end{subfigure} 
 \hfill
  \begin{subfigure}[b]{0.21\linewidth}
    \centering
      \includegraphics[width=\linewidth,keepaspectratio]{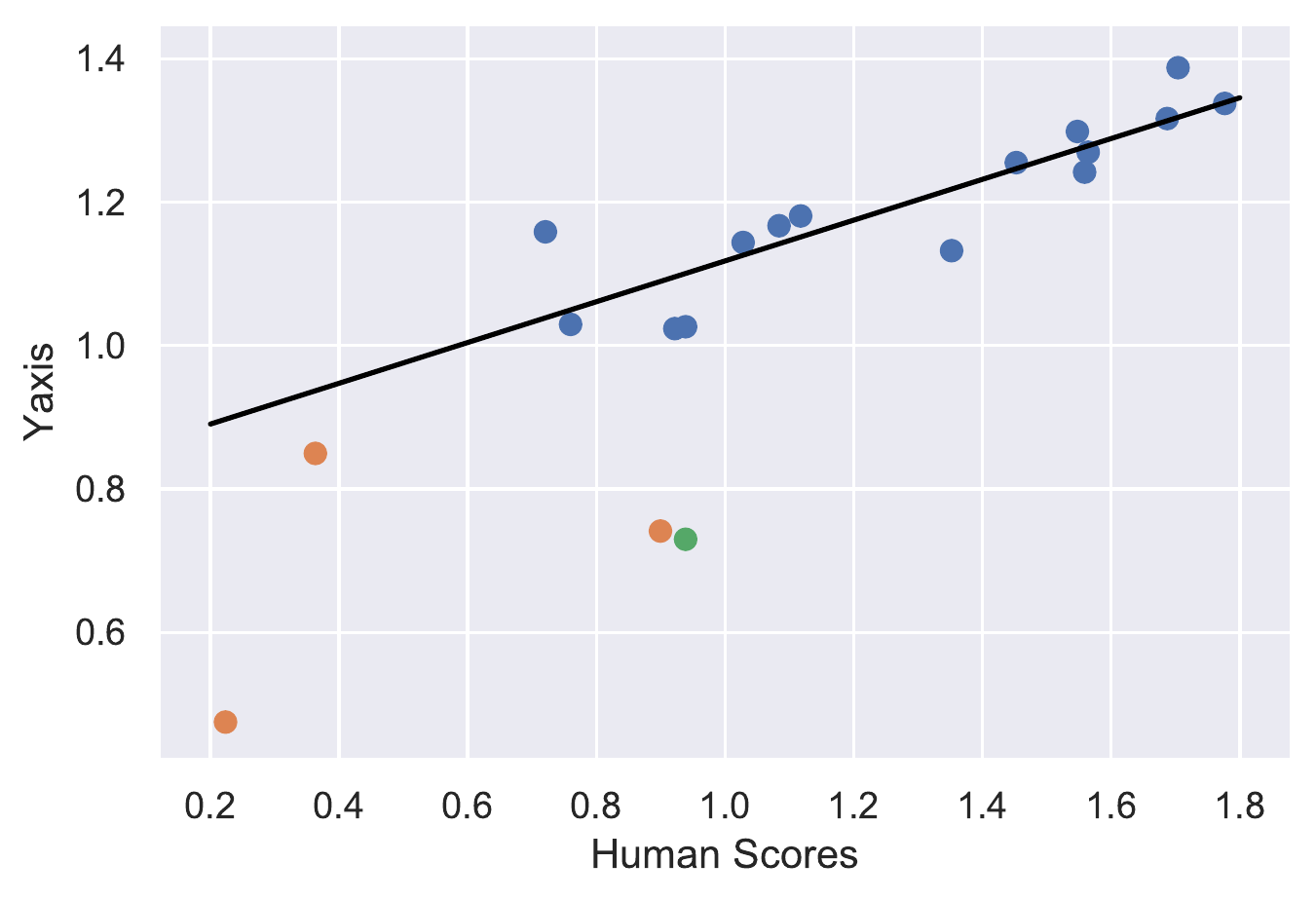}
    \caption{CNN (Human)}
    \end{subfigure} 
     \hfill
  \begin{subfigure}[b]{0.21\linewidth}
    \centering
      \includegraphics[width=\linewidth,keepaspectratio]{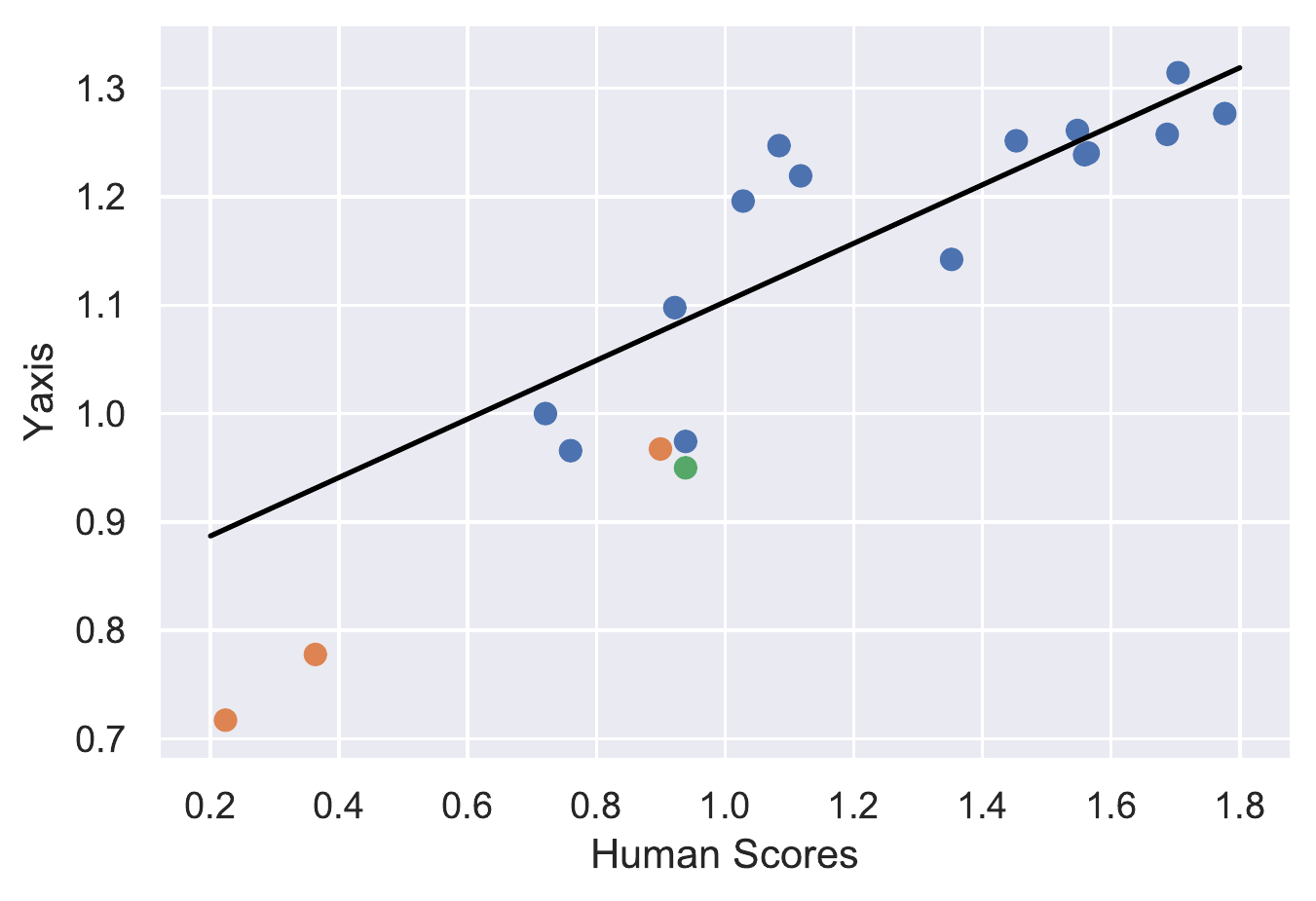}
    \caption{BERT}
    \end{subfigure} 
     \hfill
  \begin{subfigure}[b]{0.21\linewidth}
    \centering
      \includegraphics[width=\linewidth,keepaspectratio]{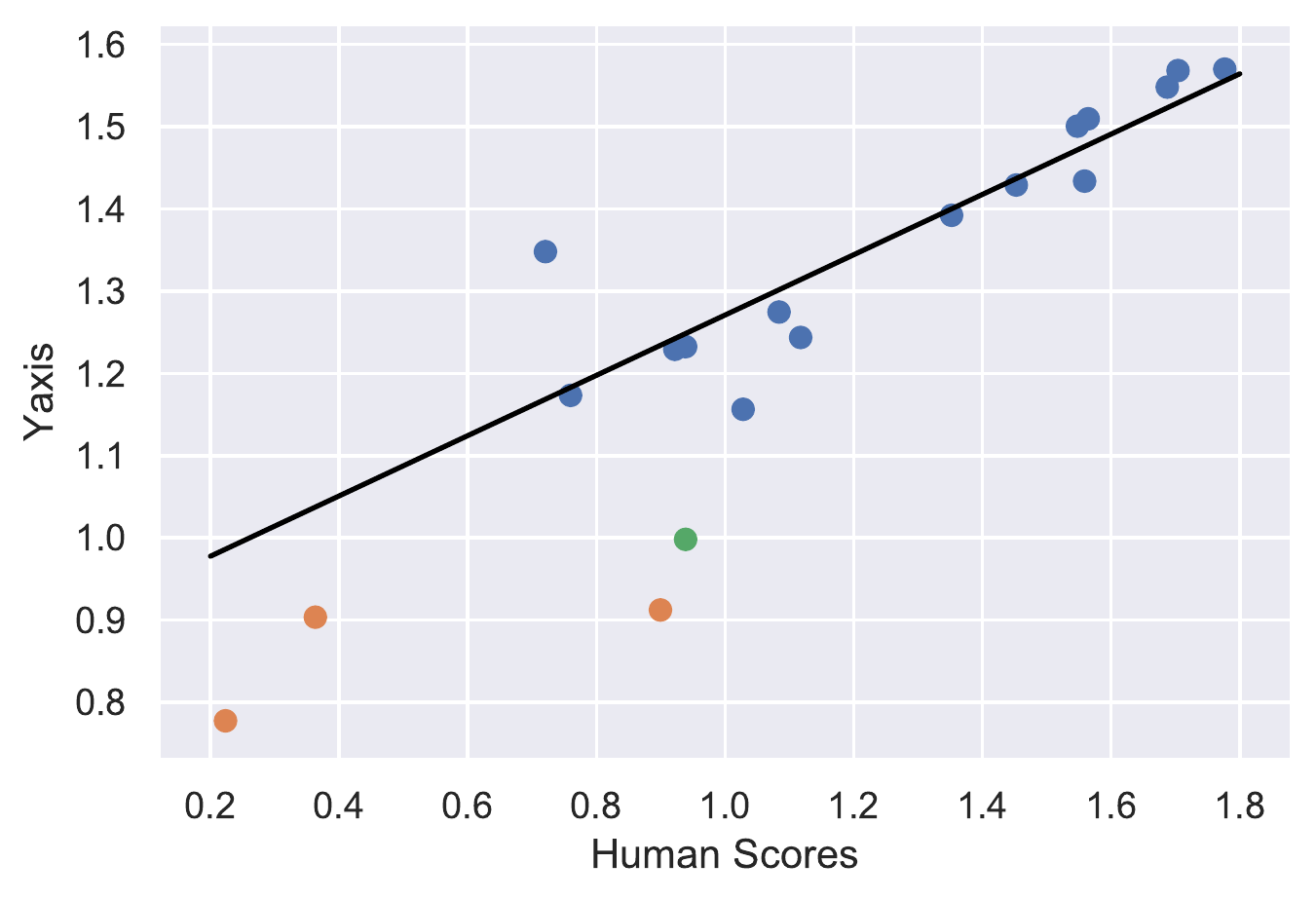}
    \caption{Longformer}
    \end{subfigure} 
 \hfill
    
    \caption{Scatter plots and best fitted lines on abstractive systems. Blue = abstractive systems, Orange = extractive systems, Green = \texttt{bart-large-cnn} system.}
    \label{fig:scatter_plot_corpus_level}    
\end{figure*}

Compared to existing data such as SummEval or RealSum, the podcast summarization task is more abstractive, and its document length is about 10 times longer. Hence, we benchmark automatic assessment methods discussed in Section \ref{section:assessment_method_setup}. The results are presented in Table \ref{tab:correlation_results} and Fig. \ref{fig:scatter_plot_corpus_level}.

% \subsubsection{System-level correlation}
\vspace{0.5em}
\noindent \textbf{Unsupervised with Reference.} The methods achieve a \textit{high} correlation. Due to the references being abstractive, ROUGE and TripleMatching with reference generally yields higher scores for abstractive systems as shown in Fig. \ref{fig:rougeL_ref} and \ref{fig:triple_ref}. 

\vspace{0.5em}
\noindent \textbf{Unsupervised with Document.} Not only these methods show a \textit{low} correlation, their correlation with human judgements is negative when including both extractive and abstractive systems. As shown in Fig. \ref{fig:rougeL_doc} to Fig. \ref{fig:entail_4096}, these methods give overly high scores to extractive systems. The summary of an extractive system by default has a high lexical overlap with the document, suggesting that although question answering (QA) and entailment approaches are not designed to directly rely on a lexical overlap, they appear to give a high score for the summary with a high lexical overlap. 

Another point is that when the input document is much longer (e.g. 6375 for podcast transcript in average) than the limit of a model (e.g. 512 for BERT), the entailment system is poor, but this can be mitigated by using a base entailment model with a larger limit such as Longformer. For the question-answering approach, we observe that swapping the question answering model from BERT to Longformer does not show an improvement. This is likely because the question answering model is trained on SQuAD2.0 data, where most answers are within BERT's length limit.

\vspace{0.5em}
\noindent \textbf{Supervised with Document.} First, a baseline CNN model is trained in a weakly supervised fashion using ROUGE-L($\mathbf{y}$,$\mathbf{y}^*$) as the target. We show that this weakly supervised approach yields a considerably higher correlation than unsupervised approaches, and it is able to learn not to score extractive systems too high. Second, we show that supervised training yields models with the highest correlation among the approaches without reference, and a correlation similar to that of ROUGE-L($\mathbf{y}$,$\mathbf{y}^*$) can be achieved. Next observation is when comparing supervised BERT and supervised Longformer. Both systems take concatenated [$\mathbf{x};\mathbf{y}$] with $\mathbf{x}$ being truncated first for long inputs. The fact that these two systems achieve a similar performance level suggests that the systems may learn to use the signal only from $\mathbf{y}$, i.e. on fluency/coherence aspect rather than the informativeness aspect.
% Unsupervised$\oplus$Doc approaches yields too high score for extractive systems. Our dataset demonstrates 

% \subsubsection{Summary-level correlation}
% Similar to other datasets, we observe a \textit{low} correlation at summary-level compared to system-level. 

\begin{table*}[!t]
  \centering
  \scalebox{0.92}{
  \begin{tabular}{r|cc|c|cc|cc}
    \toprule
    \multirow{2}{*}{Method} &\multicolumn{2}{c}{Against} &\multirow{2}{*}{Type}  &\multicolumn{2}{c}{System-level} &\multicolumn{2}{c}{Summary-level}  \\
    &Ref &Doc &     &Inc. &Exc. &Inc. &Exc. \\
    \midrule
    ROUGE-L ($\mathbf{y}$,$\mathbf{y}^*$)        &\cmark & &Unsupervised &0.905 &0.864 &0.350 &0.246 \\
    TripleMatching ($\mathbf{y}$,$\mathbf{y}^*$) &\cmark & &Unsupervised &0.838 &0.746 &0.079 &0.052   \\
    \midrule
    ROUGE-L ($\mathbf{y}$,$\mathbf{x}$) & &\cmark &Unsupervised        &-0.200 &0.364 &-0.036 &0.250 \\
    TripleMatching ($\mathbf{y}$,$\mathbf{x}$) & &\cmark &Unsupervised &-0.159 &0.453 &-0.123  &0.143   \\
    QA approach [B-512] & &\cmark &Unsupervised        &-0.112 &0.517 &-0.045 &0.123 \\
    QA approach [L-4096] & &\cmark &Unsupervised       &-0.115 &0.503 &-0.071 &0.118 \\
    Entailment [B-512] & &\cmark &Unsupervised         &0.356 &0.114 &0.102  &0.021   \\
    Entailment [L-4096] & &\cmark &Unsupervised        &-0.192 &0.392 &-0.105 &-0.059 \\
    % \midrule
    % \multicolumn{5}{c}{Supervised Without Reference Summary} \\
    \midrule
    CNN model  & &\cmark &Weakly Supervised  &0.728 &0.563 &0.171 &0.019 \\
    CNN model  & &\cmark &Supervised         &0.901 &0.902 &0.299 &0.183 \\
    BERT model  & &\cmark &Supervised        &0.905 &0.869 &0.237 &0.156 \\
    Longformer model  & &\cmark &Supervised  &0.909 &0.896 &0.278 &0.196 \\
    \bottomrule
  \end{tabular}
  }
  \caption{Spearman correlation (19 systems -- excluding creator description). Inc./Exc. = Including/Excluding extractive summaries. Pearson correlation results are provided in Appendix \ref{section:more_results1}.}
  \label{tab:correlation_results}
\end{table*}

\section{Assessment Method for Data Selection}
\label{section:assessment_for_selection}
\subsection{Absolute Score Prediction}
Another use case of summary assessment is to predict the quality on an \textit{absolute} scale. On the podcast data, a direct application is to select appropriate document-description pairs for training summarization models (discussed in Section \ref{section:sumamry_generation}). 

\subsubsection*{Baselines: Supervised Approach}
Because methods such as ROUGE, QA, or entailment do not predict a score on an absolute scale, they are not applicable. Hence, we focus on the performance of supervised approaches. We perform a 5-fold cross validation training. In Table \ref{tab:podcast_assessment_baseline}, we show the correlation against human judgements and RMSE when computed on all test samples. Despite a similar correlation at summary-level and sentence-level, the CNN model achieves the highest correlation as well as lowest variance in performance when evaluating on all test samples. 

\begin{table}[!ht]
  \centering
  \begin{tabular}{r|cc}
    \toprule
    Model      &Spearman ($\uparrow$)   &RMSE ($\downarrow$) \\
    \midrule
    CNN         &0.431{\small$\pm$0.005}  &0.884{\small$\pm$0.003} \\ 
    BERT        &0.353{\small$\pm$0.061}  &0.909{\small$\pm$0.024} \\
    Longformer  &0.397{\small$\pm$0.041}  &0.900{\small$\pm$0.019} \\
    \bottomrule
  \end{tabular}
  \caption{Absolute score prediction baselines}
  \label{tab:podcast_assessment_baseline}
\end{table}

Next, we investigate the impact of pre-training CNN with negative samples as done in \citet{bao2020end}. For pre-training, we use the CNNDM dataset where we assign 1.0 to real summaries and 0.0 to randomly selected summaries. When using the pre-trained model on podcast, the prediction is scaled up by $\times$3.0. Shown in Table \ref{tab:pretraining_cnn}, pre-trained and fine-tuned models perform worse than the vanilla model. We found that the mean prediction of pre-trained model 2.75, which is close to 3.0, suggesting that the negative sampling task is too different from the podcast task.

\begin{table}[!ht]
\tabcolsep=0.08cm
  \centering
  \begin{tabular}{r|cc}
    \toprule
    CNN Model      &Spearman ($\uparrow$)   &RMSE ($\downarrow$) \\
    \midrule
    Trained      &0.431{\small$\pm$0.005}  &0.884{\small$\pm$0.003} \\
    Pre-trained  &-0.005   &1.954   \\
    + Fine-tuned   &0.400{\small$\pm$0.006}  &0.915{\small$\pm$0.005} \\
    \bottomrule
  \end{tabular}
  \caption{Impact of pre-training.}
  \label{tab:pretraining_cnn}
\end{table}

\vspace{-0.5em}
\subsubsection*{Seen v.s. Unseen Data}
So far, we have performed cross-validation training where samples are \textit{all-shuffled}. Here, we investigate other scenarios, including: (i) when a system is held-out entirely such that no summaries from a particular system are seen at training; (ii) when some documents are held-out. Again, we train each configuration 5 times. In Table \ref{tab:seen_unseen_results}, the results show that RMSE is the highest when the \textit{creator description} set (R1) is held-out. Note that using an extractive system such as E1 is expected to yield a low correlation because most extractive summaries are graded either just fair or bad (83\% for E1, 95\% for E2, and 97\% for E3), but their RMSE values are not the worst. Next, when there are unseen documents at inference time (e.g. held-out documents), the performance is also worse than all-shuffled.

\begin{table}[h!]
\tabcolsep=0.09cm
  \centering
  \begin{tabular}{r|c|cc}
    \toprule
    n-fold &Held-out &Spearman ($\uparrow$)   &RMSE ($\downarrow$)  \\
    \midrule
    all-shuffled     &random        &0.431{\small$\pm$0.005}  &0.884{\small$\pm$0.003} \\ 
    \midrule
    system &{E1}    &0.233{\small$\pm$0.034} &0.780{\small$\pm$0.036} \\
    system &{E3}    &0.129{\small$\pm$0.055} &0.878{\small$\pm$0.079} \\
    system &{A7}   &0.473{\small$\pm$0.062} &0.968{\small$\pm$0.053}  \\
    system &{A12}  &0.540{\small$\pm$0.023} &0.958{\small$\pm$0.020}  \\
    system &{A16}  &0.434{\small$\pm$0.043} &0.888{\small$\pm$0.022}  \\
    system &{R1}    &0.245{\small$\pm$0.049} &1.035{\small$\pm$0.027} \\
    \midrule
    document  &document &0.242{\small$\pm$0.044}  &0.964{\small$\pm$0.057} \\
    \bottomrule
  \end{tabular}
  \caption{Different ways of held-out splits.}
  \label{tab:seen_unseen_results}
\end{table}

The results in Table \ref{tab:seen_unseen_results} motivate us to further investigate the scenario where there are unseen document and creator description pairs. Hence, we use all of 3580 summary assessment examples as train/valid sets (80\%/20\%), and we make use of 150 document-description pairs as the test set.\footnote{Spotify released 150 documents/episodes in the initial phase of TREC2020, and we call this set \textit{test150}.}

We found that this unseen scenario appears very challenging for the model. 22 out of 50 training runs\footnote{Each run is different by a train/valid data shuffle.} have a negative correlation on test150, and the average correlation of all 50 runs is close to zero at 0.011 as shown in Table \ref{tab:assessment_result}. 

\begin{table}[!ht]
  \centering
  \begin{tabular}{r|cc}
    \toprule
    Method  &Spearman ($\uparrow$) &RMSE ($\downarrow$)  \\
    \midrule
    SingleModel &0.011{\small$\pm$0.090} &1.100{\small$\pm$0.040} \\
    Ensemble &0.109 &1.034\\
    \bottomrule
  \end{tabular}
  \caption{Results on unseen doc-description (test150).}
  \label{tab:assessment_result}
\end{table}

\subsubsection*{Ensemble Performance and Uncertainty}
\begin{figure}[!ht]
  \begin{subfigure}[b]{0.45\linewidth}
    \centering
      \includegraphics[width=\linewidth,keepaspectratio]{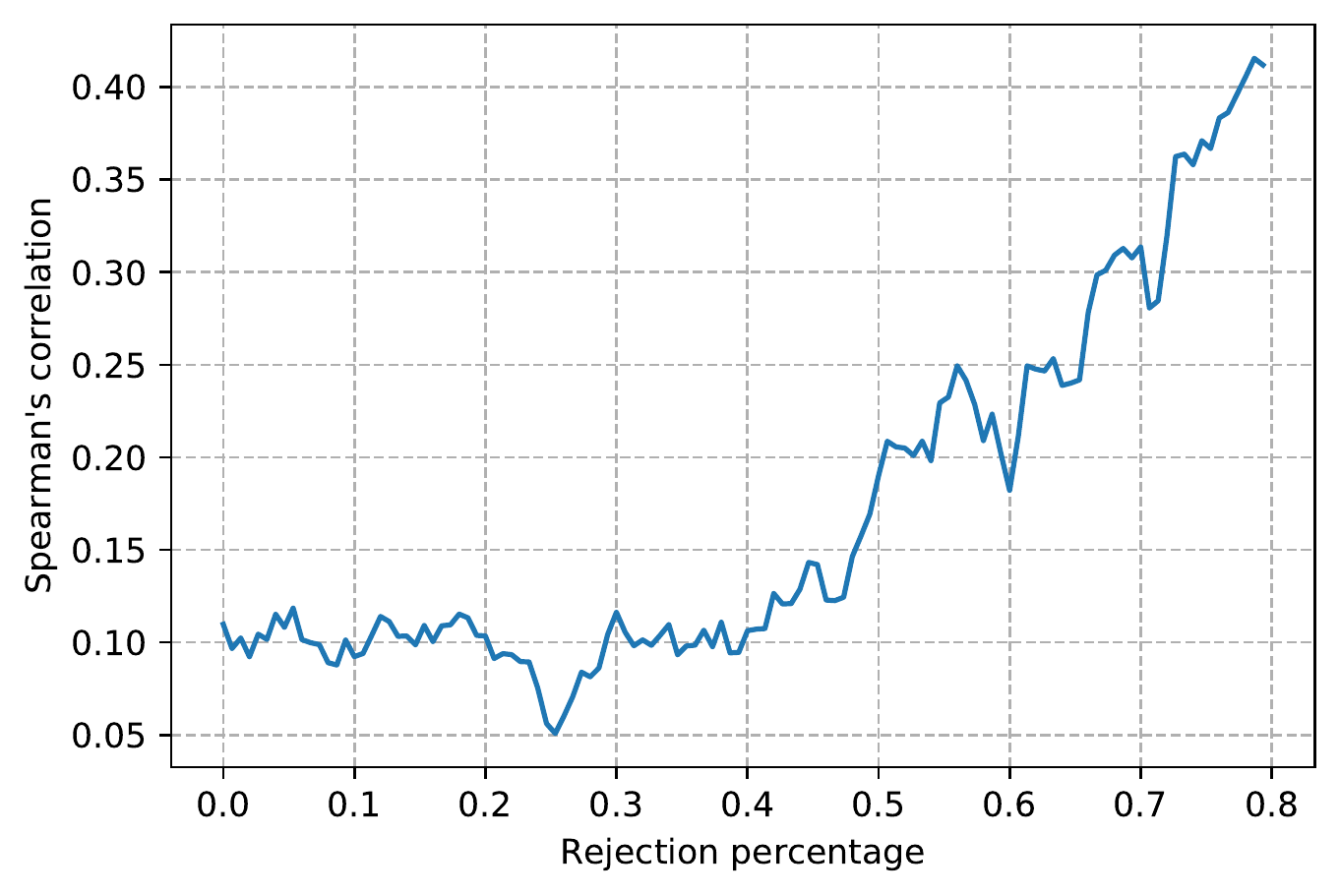}
    \caption{$\rho$ and Uncertainty}
    \end{subfigure}    
        \hfill
  \begin{subfigure}[b]{0.45\linewidth}
    \centering
      \includegraphics[width=\linewidth,keepaspectratio]{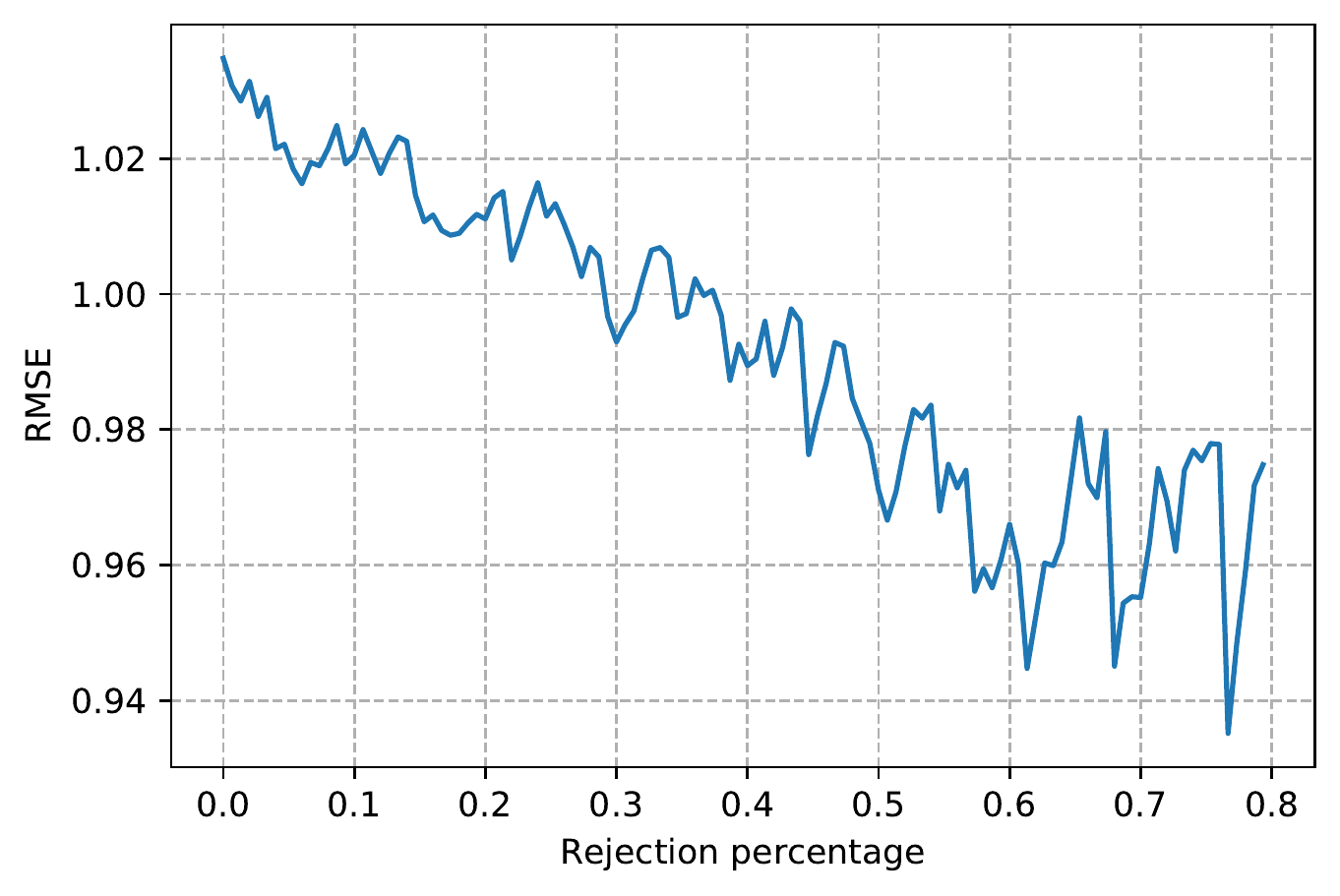}
    \caption{RMSE and Uncertainty}
    \end{subfigure}
    \caption{Uncertainty Results on test150..}
    \label{fig:uncertainty}    
\end{figure}

% \begin{figure}[!ht]
% \centering
% \includegraphics[width=0.6\linewidth,keepaspectratio]{fig/ensemble_toyset.pdf}
% \caption{Scatter plot ($\rho=0.109)$}
% \label{fig:scatter_plot_ensemble20}
% \end{figure}
To achieve the best performance, we use an ensemble by averaging the predictions of the single models. The ensemble achieves 0.109 in Spearman correlation on test150. In addition to the performance gain, the ensemble allows us to investigate uncertainty. Initial uncertainty results in Fig. \ref{fig:uncertainty} show that when the models agree the predictions are more reliable than when they are not. This suggests that uncertainty could further help the data selection task for future work.

\subsection{Summary Generation Training}
\label{section:sumamry_generation}
% In summary generation experiments, we refer to the entire Spotify Podcast dataset of more than 100,000 episodes \cite{clifton-etal-2020-100000} rather than podcast summary assessment data.
Because the podcast summarization dataset does \textit{not} have perfect or gold summaries for training and evaluating summary generation models, previous work filtered training set, down from 105k to 60k examples, using simple heuristics\footnote{More information in Appendix \ref{appendix:8questions}.} \cite{manakul2020cued_speech}. This filtered set is called \textit{brass} set. In this work, we investigate if assessment models can be used to perform data selection.

We use the ensemble system (in Table \ref{tab:assessment_result}) for selecting document-description training examples. We run the system on the entire podcast summarization training set of 105k examples. We create \textit{top} set where we keep 60k examples of the highest assessment scores and \textit{bottom} set where we keep 60k examples of the lowest scores.

We train BART on each training set in Table \ref{tab:bart_train_set} using the best configuration described in \citet{manakul2021_longspan}: ORC-pad-rand is applied to select sentences at training time, and model-based MCS is applied at inference time. Note that we keep the same valid/test sets.  
\begin{table}[!ht]
  \tabcolsep=1.3mm
  \centering
  \begin{tabular}{r|c|l}
    \toprule
    Train-set  &Size &Description \\
    \midrule
    All     &105k      &All training examples \\
    Brass   &60k       &Selection based on heuristics \\
    Top     &60k       &Examples of highest score \\
    Bottom  &60k       &Examples of lowest score \\
    \bottomrule
  \end{tabular}
  \caption{Training sets for summarization systems.}
  \label{tab:bart_train_set}
\end{table}

\subsubsection*{Impact on Assessment System Score}
\begin{figure}[!ht]
\centering
\includegraphics[width=0.575\linewidth,keepaspectratio]{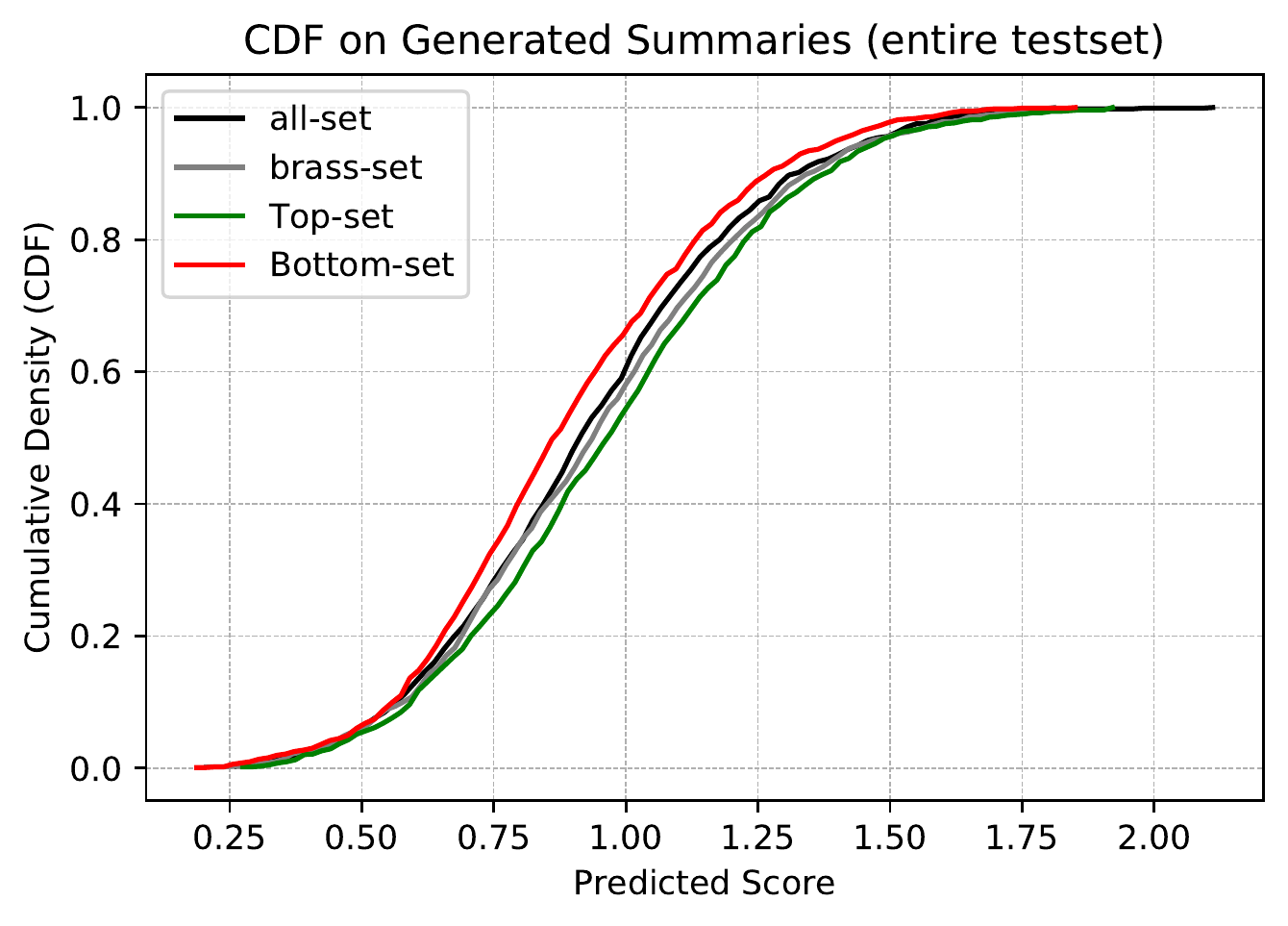}
\caption{Cumulative density plot of the assessment scores on testset.}
\label{fig:testset_predict_score_dist}
\end{figure}

We generate summaries of the summarization testset (1027 examples) using BART trained with different training sets. Then, we predict the summary quality score. Table \ref{tab:predicted_scores} and Fig. \ref{fig:testset_predict_score_dist} support that using assessment model to select training set is able to shift the summarization model towards generating summaries that either have a \textit{higher} or \textit{lower} assessment score at inference time. Therefore, this simple training data selection via assessment method can guide the summarization model. 

\begin{table}[!ht]
  \tabcolsep=1.5mm
  \centering
  \begin{tabular}{r|cc}
    \toprule
    \multirow{2}{*}{Summarization Model}   &\multicolumn{2}{c}{Average Score} \\
             &Train-set   &Test-set \\
    \midrule
    All         &1.053   &0.941 \\
    Brass       &1.083   &0.956 \\
    Top         &1.236   &0.982 \\
    Bottom      &0.867   &0.900 \\
    \bottomrule
  \end{tabular}
  \caption{Assessment score in range [0.0, 3.0] predicted by our ensemble system on testset.}
  \label{tab:predicted_scores}
\end{table}

% \vspace{-1em}
\subsubsection*{Impact on ROUGE}
Despite the generated summaries of BART trained on \textit{top}-score set obtaining the highest assessment system score, the performance measured by ROUGE (in Table \ref{tab:training_results}) does not show an improvement over BART trained on all/brass/bottom sets.  

It should be noted that the testset set contains all EGFB grades, and a higher ROUGE score may only indicate that generated summaries are lexically closer to the summaries in the testset. Figure \ref{fig:rougeL_ref} also reveals that the correlation between ROUGE and human judgement is \textit{low} or even negative when considering only top systems, e.g. system-level $\rho$ = -0.28 for top-7 systems. We suggest that more attention is required when comparing high performing systems using ROUGE.

\begin{table}[!ht]
  \tabcolsep=1.6mm
  \centering
  \begin{tabular}{r|ccc}
    \toprule
    Summarization Model &R1 &R2 &RL  \\
    \midrule
    All       &28.46 &11.19 &20.08 \\
    Brass     &27.28 &9.82 &19.00 \\
    Top       &27.22 &9.81 &18.87 \\
    Bottom    &27.52 &10.43 &19.37 \\
    \bottomrule
  \end{tabular}
  \caption{Summarization system development results.}
  \label{tab:training_results}
\end{table}

\vspace{-0.5em}
\section{Conclusion}
This work has assembled and released a new resource for summary assessment. The corpus is unique in that the data consists of podcast episodes, instead of news articles which have received more attention. This corpus has two interesting aspects that the documents are long, and there is a challenge in applying summary assessment methods to improvement the summary generation task. We provide benchmark results of existing assessment methods on this new corpus as a baseline for future work. In addition, we apply model-based supervised assessment methods to select data for the generation task, and we provide initial results and insights based on the new corpus. 
\newpage
\bibliography{anthology, custom}

\begin{thebibliography}{45}
\expandafter\ifx\csname natexlab\endcsname\relax\def\natexlab#1{#1}\fi

\bibitem[{Angeli et~al.(2015)Angeli, Johnson~Premkumar, and
  Manning}]{angeli-etal-2015-leveraging}
Gabor Angeli, Melvin~Jose Johnson~Premkumar, and Christopher~D. Manning. 2015.
\newblock \href {https://doi.org/10.3115/v1/P15-1034} {Leveraging linguistic
  structure for open domain information extraction}.
\newblock In \emph{Proceedings of the 53rd Annual Meeting of the Association
  for Computational Linguistics and the 7th International Joint Conference on
  Natural Language Processing (Volume 1: Long Papers)}, pages 344--354,
  Beijing, China. Association for Computational Linguistics.

\bibitem[{Bao et~al.(2020)Bao, Li, Luo, Chen, Yang, He, and Qiu}]{bao2020end}
Forrest~Sheng Bao, Hebi Li, Ge~Luo, Cen Chen, Yinfei Yang, Youbiao He, and
  Minghui Qiu. 2020.
\newblock End-to-end semantics-based summary quality assessment for
  single-document summarization.
\newblock \emph{arXiv preprint arXiv:2005.06377}.

\bibitem[{Beltagy et~al.(2020)Beltagy, Peters, and
  Cohan}]{beltagy2020longformer}
Iz~Beltagy, Matthew~E Peters, and Arman Cohan. 2020.
\newblock \href {https://arxiv.org/abs/2004.05150} {Longformer: The
  long-document transformer}.
\newblock \emph{arXiv preprint arXiv:2004.05150}.

\bibitem[{Bhandari et~al.(2020)Bhandari, Gour, Ashfaq, Liu, and
  Neubig}]{bhandari-etal-2020-evaluating}
Manik Bhandari, Pranav~Narayan Gour, Atabak Ashfaq, Pengfei Liu, and Graham
  Neubig. 2020.
\newblock \href {https://doi.org/10.18653/v1/2020.emnlp-main.751}
  {Re-evaluating evaluation in text summarization}.
\newblock In \emph{Proceedings of the 2020 Conference on Empirical Methods in
  Natural Language Processing (EMNLP)}, pages 9347--9359, Online. Association
  for Computational Linguistics.

\bibitem[{Clifton et~al.(2020)Clifton, Reddy, Yu, Pappu, Rezapour, Bonab,
  Eskevich, Jones, Karlgren, Carterette, and Jones}]{clifton-etal-2020-100000}
Ann Clifton, Sravana Reddy, Yongze Yu, Aasish Pappu, Rezvaneh Rezapour, Hamed
  Bonab, Maria Eskevich, Gareth Jones, Jussi Karlgren, Ben Carterette, and
  Rosie Jones. 2020.
\newblock \href {https://doi.org/10.18653/v1/2020.coling-main.519} {100,000
  podcasts: A spoken {E}nglish document corpus}.
\newblock In \emph{Proceedings of the 28th International Conference on
  Computational Linguistics}, pages 5903--5917, Barcelona, Spain (Online).
  International Committee on Computational Linguistics.

\bibitem[{Dang and Owczarzak(2008)}]{dang2008OverviewOT}
Hoa~Trang Dang and Karolina Owczarzak. 2008.
\newblock Overview of the tac 2008 update summarization task.
\newblock \emph{TAC}.

\bibitem[{Dang and Owczarzak(2009)}]{dang2009OverviewOT}
Hoa~Trang Dang and Karolina Owczarzak. 2009.
\newblock Overview of the tac 2009 summarization track.
\newblock \emph{TAC}.

\bibitem[{Deutsch et~al.(2021)Deutsch, Bedrax-Weiss, and
  Roth}]{deutsch-etal-2021-towards}
Daniel Deutsch, Tania Bedrax-Weiss, and Dan Roth. 2021.
\newblock \href {https://doi.org/10.1162/tacl_a_00397} {Towards
  question-answering as an automatic metric for evaluating the content quality
  of a summary}.
\newblock \emph{Transactions of the Association for Computational Linguistics},
  9:774--789.

\bibitem[{Devlin et~al.(2019)Devlin, Chang, Lee, and
  Toutanova}]{devlin-etal-2019-bert}
Jacob Devlin, Ming-Wei Chang, Kenton Lee, and Kristina Toutanova. 2019.
\newblock \href {https://doi.org/10.18653/v1/N19-1423} {{BERT}: Pre-training of
  deep bidirectional transformers for language understanding}.
\newblock In \emph{Proceedings of the 2019 Conference of the North {A}merican
  Chapter of the Association for Computational Linguistics: Human Language
  Technologies, Volume 1 (Long and Short Papers)}, pages 4171--4186,
  Minneapolis, Minnesota. Association for Computational Linguistics.

\bibitem[{Durmus et~al.(2020)Durmus, He, and Diab}]{durmus-etal-2020-feqa}
Esin Durmus, He~He, and Mona Diab. 2020.
\newblock \href {https://doi.org/10.18653/v1/2020.acl-main.454} {{FEQA}: A
  question answering evaluation framework for faithfulness assessment in
  abstractive summarization}.
\newblock In \emph{Proceedings of the 58th Annual Meeting of the Association
  for Computational Linguistics}, pages 5055--5070, Online. Association for
  Computational Linguistics.

\bibitem[{Fabbri et~al.(2021)Fabbri, Kry{\'s}ci{\'n}ski, McCann, Xiong, Socher,
  and Radev}]{fabbri2021summeval}
Alexander~R Fabbri, Wojciech Kry{\'s}ci{\'n}ski, Bryan McCann, Caiming Xiong,
  Richard Socher, and Dragomir Radev. 2021.
\newblock Summeval: Re-evaluating summarization evaluation.
\newblock \emph{Transactions of the Association for Computational Linguistics},
  9:391--409.

\bibitem[{Goodrich et~al.(2019)Goodrich, Rao, Liu, and Saleh}]{goodrich_triple}
Ben Goodrich, Vinay Rao, Peter~J. Liu, and Mohammad Saleh. 2019.
\newblock \href {https://doi.org/10.1145/3292500.3330955} {Assessing the
  factual accuracy of generated text}.
\newblock In \emph{Proceedings of the 25th ACM SIGKDD International Conference
  on Knowledge Discovery; Data Mining}, KDD '19, page 166–175, New York, NY,
  USA. Association for Computing Machinery.

\bibitem[{Grusky et~al.(2018)Grusky, Naaman, and
  Artzi}]{grusky-etal-2018-newsroom}
Max Grusky, Mor Naaman, and Yoav Artzi. 2018.
\newblock \href {https://doi.org/10.18653/v1/N18-1065} {{N}ewsroom: A dataset
  of 1.3 million summaries with diverse extractive strategies}.
\newblock In \emph{Proceedings of the 2018 Conference of the North {A}merican
  Chapter of the Association for Computational Linguistics: Human Language
  Technologies, Volume 1 (Long Papers)}, pages 708--719, New Orleans,
  Louisiana. Association for Computational Linguistics.

\bibitem[{He et~al.(2016)He, Zhang, Ren, and Sun}]{he2016deep}
Kaiming He, Xiangyu Zhang, Shaoqing Ren, and Jian Sun. 2016.
\newblock Deep residual learning for image recognition.
\newblock In \emph{Proceedings of the IEEE conference on computer vision and
  pattern recognition}, pages 770--778.

\bibitem[{Hermann et~al.(2015)Hermann, Kocisk{\'{y}}, Grefenstette, Espeholt,
  Kay, Suleyman, and Blunsom}]{hermann2015teaching}
Karl~Moritz Hermann, Tom{\'{a}}s Kocisk{\'{y}}, Edward Grefenstette, Lasse
  Espeholt, Will Kay, Mustafa Suleyman, and Phil Blunsom. 2015.
\newblock \href
  {https://proceedings.neurips.cc/paper/2015/hash/afdec7005cc9f14302cd0474fd0f3c96-Abstract.html}
  {Teaching machines to read and comprehend}.
\newblock In \emph{Advances in Neural Information Processing Systems 28: Annual
  Conference on Neural Information Processing Systems 2015, December 7-12,
  2015, Montreal, Quebec, Canada}, pages 1693--1701.

\bibitem[{Jones et~al.(2020)Jones, Carterette, Clifton, Eskevich, Jones,
  Karlgren, Pappu, Reddy, and Yu}]{jones_trec2020}
Rosie Jones, Ben Carterette, Ann Clifton, Maria Eskevich, Gareth J.~F. Jones,
  Jussi Karlgren, Aasish Pappu, Sravana Reddy, and Yongze Yu. 2020.
\newblock Trec 2020 podcasts track overview.
\newblock In \emph{The 29th Text Retrieval Conference (TREC) notebook}.

\bibitem[{Karlbom and Clifton(2020)}]{hk_uu_trec}
Hannes Karlbom and Ann Clifton. 2020.
\newblock Abstract podcast summarization using {BART} with longformer
  attention.
\newblock In \emph{Proceedings of the 29th {T}ext {RE}trieval Conference
  ({TREC})}.

\bibitem[{Koto et~al.(2022)Koto, Baldwin, and Lau}]{koto2022ffci}
Fajri Koto, Timothy Baldwin, and Jey~Han Lau. 2022.
\newblock Ffci: A framework for interpretable automatic evaluation of
  summarization.
\newblock \emph{Journal of Artificial Intelligence Research}, 73:1553--1607.

\bibitem[{Kryscinski et~al.(2020)Kryscinski, McCann, Xiong, and
  Socher}]{kryscinski-etal-2020-evaluating}
Wojciech Kryscinski, Bryan McCann, Caiming Xiong, and Richard Socher. 2020.
\newblock \href {https://doi.org/10.18653/v1/2020.emnlp-main.750} {Evaluating
  the factual consistency of abstractive text summarization}.
\newblock In \emph{Proceedings of the 2020 Conference on Empirical Methods in
  Natural Language Processing (EMNLP)}, pages 9332--9346, Online. Association
  for Computational Linguistics.

\bibitem[{Lewis et~al.(2020)Lewis, Liu, Goyal, Ghazvininejad, Mohamed, Levy,
  Stoyanov, and Zettlemoyer}]{lewis-etal-2020-bart}
Mike Lewis, Yinhan Liu, Naman Goyal, Marjan Ghazvininejad, Abdelrahman Mohamed,
  Omer Levy, Veselin Stoyanov, and Luke Zettlemoyer. 2020.
\newblock \href {https://doi.org/10.18653/v1/2020.acl-main.703} {{BART}:
  Denoising sequence-to-sequence pre-training for natural language generation,
  translation, and comprehension}.
\newblock In \emph{Proceedings of the 58th Annual Meeting of the Association
  for Computational Linguistics}, pages 7871--7880, Online. Association for
  Computational Linguistics.

\bibitem[{Lin(2004)}]{lin-2004-rouge}
Chin-Yew Lin. 2004.
\newblock \href {https://www.aclweb.org/anthology/W04-1013} {{ROUGE}: A package
  for automatic evaluation of summaries}.
\newblock In \emph{Text Summarization Branches Out}, pages 74--81, Barcelona,
  Spain. Association for Computational Linguistics.

\bibitem[{Manakul and Gales(2020)}]{manakul2020cued_speech}
Potsawee Manakul and Mark Gales. 2020.
\newblock {CUED}\_speech at {TREC} 2020 {P}odcast {S}ummarisation {T}rack.
\newblock \emph{arXiv preprint arXiv:2012.02535}.

\bibitem[{Manakul and Gales(2021)}]{manakul2021_longspan}
Potsawee Manakul and Mark Gales. 2021.
\newblock \href {https://doi.org/10.18653/v1/2021.acl-long.470} {Long-span
  summarization via local attention and content selection}.
\newblock In \emph{Proceedings of the 59th Annual Meeting of the Association
  for Computational Linguistics and the 11th International Joint Conference on
  Natural Language Processing (Volume 1: Long Papers)}, pages 6026--6041,
  Online. Association for Computational Linguistics.

\bibitem[{Maynez et~al.(2020)Maynez, Narayan, Bohnet, and
  McDonald}]{maynez-etal-2020-faithfulness}
Joshua Maynez, Shashi Narayan, Bernd Bohnet, and Ryan McDonald. 2020.
\newblock \href {https://doi.org/10.18653/v1/2020.acl-main.173} {On
  faithfulness and factuality in abstractive summarization}.
\newblock In \emph{Proceedings of the 58th Annual Meeting of the Association
  for Computational Linguistics}, pages 1906--1919, Online. Association for
  Computational Linguistics.

\bibitem[{Mihalcea and Tarau(2004)}]{mihalcea-tarau-2004-textrank}
Rada Mihalcea and Paul Tarau. 2004.
\newblock \href {https://www.aclweb.org/anthology/W04-3252} {{T}ext{R}ank:
  Bringing order into text}.
\newblock In \emph{Proceedings of the 2004 Conference on Empirical Methods in
  Natural Language Processing}, pages 404--411, Barcelona, Spain. Association
  for Computational Linguistics.

\bibitem[{Narayan et~al.(2018)Narayan, Cohen, and
  Lapata}]{narayan-etal-2018-dont}
Shashi Narayan, Shay~B. Cohen, and Mirella Lapata. 2018.
\newblock \href {https://doi.org/10.18653/v1/D18-1206} {Don{'}t give me the
  details, just the summary! topic-aware convolutional neural networks for
  extreme summarization}.
\newblock In \emph{Proceedings of the 2018 Conference on Empirical Methods in
  Natural Language Processing}, pages 1797--1807, Brussels, Belgium.
  Association for Computational Linguistics.

\bibitem[{Owoicho and Dalton(2020)}]{glasgow_trec}
Paul Owoicho and Jeff Dalton. 2020.
\newblock {G}lasgow {R}epresentation and {I}nformation {L}earning {L}ab
  ({GRILL}) at {TREC} 2020 {P}odcasts {T}rack.
\newblock In \emph{Proceedings of the 29th Text REtrieval Conference (TREC)}.

\bibitem[{Papineni et~al.(2002)Papineni, Roukos, Ward, and
  Zhu}]{papineni-etal-2002-bleu}
Kishore Papineni, Salim Roukos, Todd Ward, and Wei-Jing Zhu. 2002.
\newblock \href {https://doi.org/10.3115/1073083.1073135} {{B}leu: a method for
  automatic evaluation of machine translation}.
\newblock In \emph{Proceedings of the 40th Annual Meeting of the Association
  for Computational Linguistics}, pages 311--318, Philadelphia, Pennsylvania,
  USA. Association for Computational Linguistics.

\bibitem[{Raffel et~al.(2020)Raffel, Shazeer, Roberts, Lee, Narang, Matena,
  Zhou, Li, and Liu}]{raffel2020exploring}
Colin Raffel, Noam Shazeer, Adam Roberts, Katherine Lee, Sharan Narang, Michael
  Matena, Yanqi Zhou, Wei Li, and Peter~J. Liu. 2020.
\newblock \href {http://jmlr.org/papers/v21/20-074.html} {Exploring the limits
  of transfer learning with a unified text-to-text transformer}.
\newblock \emph{Journal of Machine Learning Research}, 21(140):1--67.

\bibitem[{Rei et~al.(2020)Rei, Stewart, Farinha, and
  Lavie}]{rei-etal-2020-comet}
Ricardo Rei, Craig Stewart, Ana~C Farinha, and Alon Lavie. 2020.
\newblock \href {https://doi.org/10.18653/v1/2020.emnlp-main.213} {{COMET}: A
  neural framework for {MT} evaluation}.
\newblock In \emph{Proceedings of the 2020 Conference on Empirical Methods in
  Natural Language Processing (EMNLP)}, pages 2685--2702, Online. Association
  for Computational Linguistics.

\bibitem[{Reimers and Gurevych(2019)}]{reimers-gurevych-2019-sentence}
Nils Reimers and Iryna Gurevych. 2019.
\newblock \href {https://doi.org/10.18653/v1/D19-1410} {Sentence-{BERT}:
  Sentence embeddings using {S}iamese {BERT}-networks}.
\newblock In \emph{Proceedings of the 2019 Conference on Empirical Methods in
  Natural Language Processing and the 9th International Joint Conference on
  Natural Language Processing (EMNLP-IJCNLP)}, pages 3982--3992, Hong Kong,
  China. Association for Computational Linguistics.

\bibitem[{Scialom et~al.(2021)Scialom, Dray, Lamprier, Piwowarski, Staiano,
  Wang, and Gallinari}]{scialom-etal-2021-questeval}
Thomas Scialom, Paul-Alexis Dray, Sylvain Lamprier, Benjamin Piwowarski, Jacopo
  Staiano, Alex Wang, and Patrick Gallinari. 2021.
\newblock \href {https://doi.org/10.18653/v1/2021.emnlp-main.529}
  {{Q}uest{E}val: Summarization asks for fact-based evaluation}.
\newblock In \emph{Proceedings of the 2021 Conference on Empirical Methods in
  Natural Language Processing}, pages 6594--6604, Online and Punta Cana,
  Dominican Republic. Association for Computational Linguistics.

\bibitem[{See et~al.(2017)See, Liu, and Manning}]{see-etal-2017-get}
Abigail See, Peter~J. Liu, and Christopher~D. Manning. 2017.
\newblock \href {https://doi.org/10.18653/v1/P17-1099} {Get to the point:
  Summarization with pointer-generator networks}.
\newblock In \emph{Proceedings of the 55th Annual Meeting of the Association
  for Computational Linguistics (Volume 1: Long Papers)}, pages 1073--1083,
  Vancouver, Canada. Association for Computational Linguistics.

\bibitem[{Sellam et~al.(2020)Sellam, Das, and Parikh}]{sellam-etal-2020-bleurt}
Thibault Sellam, Dipanjan Das, and Ankur Parikh. 2020.
\newblock \href {https://doi.org/10.18653/v1/2020.acl-main.704} {{BLEURT}:
  Learning robust metrics for text generation}.
\newblock In \emph{Proceedings of the 58th Annual Meeting of the Association
  for Computational Linguistics}, pages 7881--7892, Online. Association for
  Computational Linguistics.

\bibitem[{Song et~al.(2020)Song, Li, Wang, Yu, and Liu}]{song2020automatic}
Kaiqiang Song, Chen Li, Xiaoyang Wang, Dong Yu, and Fei Liu. 2020.
\newblock Automatic summarization of open-domain podcast episodes.
\newblock \emph{arXiv preprint arXiv:2011.04132}.

\bibitem[{Trischler et~al.(2017)Trischler, Wang, Yuan, Harris, Sordoni,
  Bachman, and Suleman}]{trischler-etal-2017-newsqa}
Adam Trischler, Tong Wang, Xingdi Yuan, Justin Harris, Alessandro Sordoni,
  Philip Bachman, and Kaheer Suleman. 2017.
\newblock \href {https://doi.org/10.18653/v1/W17-2623} {{N}ews{QA}: A machine
  comprehension dataset}.
\newblock In \emph{Proceedings of the 2nd Workshop on Representation Learning
  for {NLP}}, pages 191--200, Vancouver, Canada. Association for Computational
  Linguistics.

\bibitem[{Wang et~al.(2020)Wang, Cho, and Lewis}]{wang-etal-2020-asking}
Alex Wang, Kyunghyun Cho, and Mike Lewis. 2020.
\newblock \href {https://doi.org/10.18653/v1/2020.acl-main.450} {Asking and
  answering questions to evaluate the factual consistency of summaries}.
\newblock In \emph{Proceedings of the 58th Annual Meeting of the Association
  for Computational Linguistics}, pages 5008--5020, Online. Association for
  Computational Linguistics.

\bibitem[{Williams et~al.(2018)Williams, Nangia, and
  Bowman}]{williams-etal-2018-broad}
Adina Williams, Nikita Nangia, and Samuel Bowman. 2018.
\newblock \href {https://doi.org/10.18653/v1/N18-1101} {A broad-coverage
  challenge corpus for sentence understanding through inference}.
\newblock In \emph{Proceedings of the 2018 Conference of the North {A}merican
  Chapter of the Association for Computational Linguistics: Human Language
  Technologies, Volume 1 (Long Papers)}, pages 1112--1122, New Orleans,
  Louisiana. Association for Computational Linguistics.

\bibitem[{Wu et~al.(2020)Wu, Ma, Wu, Manyumwa, and
  Ji}]{wu-etal-2020-unsupervised}
Hanlu Wu, Tengfei Ma, Lingfei Wu, Tariro Manyumwa, and Shouling Ji. 2020.
\newblock \href {https://doi.org/10.18653/v1/2020.emnlp-main.294} {Unsupervised
  reference-free summary quality evaluation via contrastive learning}.
\newblock In \emph{Proceedings of the 2020 Conference on Empirical Methods in
  Natural Language Processing (EMNLP)}, pages 3612--3621, Online. Association
  for Computational Linguistics.

\bibitem[{Xia et~al.(2019)Xia, Kochmar, and Briscoe}]{xia-etal-2019-automatic}
Menglin Xia, Ekaterina Kochmar, and Ted Briscoe. 2019.
\newblock \href {https://doi.org/10.18653/v1/N19-1261} {Automatic learner
  summary assessment for reading comprehension}.
\newblock In \emph{Proceedings of the 2019 Conference of the North {A}merican
  Chapter of the Association for Computational Linguistics: Human Language
  Technologies, Volume 1 (Long and Short Papers)}, pages 2532--2542,
  Minneapolis, Minnesota. Association for Computational Linguistics.

\bibitem[{Yuan et~al.(2021)Yuan, Neubig, and Liu}]{yuan2021bartscore}
Weizhe Yuan, Graham Neubig, and Pengfei Liu. 2021.
\newblock \href
  {https://proceedings.neurips.cc/paper/2021/file/e4d2b6e6fdeca3e60e0f1a62fee3d9dd-Paper.pdf}
  {Bartscore: Evaluating generated text as text generation}.
\newblock In \emph{Advances in Neural Information Processing Systems},
  volume~34, pages 27263--27277. Curran Associates, Inc.

\bibitem[{Zhang et~al.(2020)Zhang, Zhao, Saleh, and Liu}]{zhang2020pegasus}
Jingqing Zhang, Yao Zhao, Mohammad Saleh, and Peter Liu. 2020.
\newblock Pegasus: Pre-training with extracted gap-sentences for abstractive
  summarization.
\newblock In \emph{International Conference on Machine Learning}, pages
  11328--11339. PMLR.

\bibitem[{Zhang* et~al.(2020)Zhang*, Kishore*, Wu*, Weinberger, and
  Artzi}]{BERTScore}
Tianyi Zhang*, Varsha Kishore*, Felix Wu*, Kilian~Q. Weinberger, and Yoav
  Artzi. 2020.
\newblock \href {https://openreview.net/forum?id=SkeHuCVFDr} {Bertscore:
  Evaluating text generation with bert}.
\newblock In \emph{International Conference on Learning Representations}.

\bibitem[{Zhao et~al.(2019)Zhao, Peyrard, Liu, Gao, Meyer, and
  Eger}]{zhao-etal-2019-moverscore}
Wei Zhao, Maxime Peyrard, Fei Liu, Yang Gao, Christian~M. Meyer, and Steffen
  Eger. 2019.
\newblock \href {https://doi.org/10.18653/v1/D19-1053} {{M}over{S}core: Text
  generation evaluating with contextualized embeddings and earth mover
  distance}.
\newblock In \emph{Proceedings of the 2019 Conference on Empirical Methods in
  Natural Language Processing and the 9th International Joint Conference on
  Natural Language Processing (EMNLP-IJCNLP)}, pages 563--578, Hong Kong,
  China. Association for Computational Linguistics.

\bibitem[{Zheng et~al.(2020)Zheng, Zhang, Wang, and Fan}]{zheng2020two}
Chujie Zheng, Kunpeng Zhang, Harry~Jiannan Wang, and Ling Fan. 2020.
\newblock A two-phase approach for abstractive podcast summarization.
\newblock \emph{arXiv preprint arXiv:2011.08291}.

\end{thebibliography}
\bibliographystyle{acl_natbib}

\appendix
\section{Implementation}
\label{section:appendix_implement}

\noindent \textbf{TripleMatching}: We use Standford's CoreNLP OpenIE \cite{angeli-etal-2015-leveraging} as in the information extraction module.

\vspace{0.5em}
\noindent \textbf{CNN}: The network has ResNet18 backbone \cite{he2016deep} followed by a dropout layer ($p=0.2$) and a linear layer ($1000 \rightarrow 1$). Sentence similarity grid is obtained via cosine-similarity between every pair of document sentences and summary sentences. Each similarity grid is resized to $640\times32$. The sentence representation is based on Sentence-BERT (bert-large-nli-mean-tokens).

\vspace{0.5em}
\noindent \textbf{BERT/Longformer}: For supervised training, we use pre-trained weights from HuggingFace as follows:  "bert-large-uncased" for BERT and "allenai/longformer-base-4096" for Longformer.

\vspace{0.5em}
\noindent \textbf{Supervised Training}: We use the Adam optimizer with $10^{-5}$ learning rate, and we adopt early stopping, i.e. stop training when the validation loss does not improve.

\section{More information on Data}
\label{appendix:8questions}
\noindent \textbf{8 binary attributes}: In addition to overall scores, NIST annotators also labelled 8 binary attributes (Yes/No questions) for each summary as follows \cite{jones_trec2020}: (1) names of the main people included?; (2) any additional information about the people mentioned?; (3) main topic included?; (4) format of the podcast mentioned?; (5) context on title?; (6) redundant information?; (7) good written English? (8) good start and end points?.

\vspace{0.5em}
\noindent \textbf{brass set}: \citet{jones_trec2020} filtered the entire summarization training set using three heuristics: (1) too long (>750 characters) or too short (<20 characters); (2) description too similar to other descriptions; (3) description too similar to its show description. Similarity is calculated using \texttt{sklearn}.

\section{Additional correlation results}
\label{section:more_results1}
\begin{table}[!ht]
  \tabcolsep=1.5mm
  \centering
    \scalebox{0.79}{
  \begin{tabular}{r|cc|cc}
    \toprule
    \multirow{2}{*}{Method}  &\multicolumn{2}{c}{System-lvl} &\multicolumn{2}{c}{Summary-lvl}  \\
    &Inc. &Exc. &Inc. &Exc. \\
    \midrule
    R-L ($\mathbf{y}$,$\mathbf{y}^*$)       &0.919 &0.868 &0.326 &0.226 \\
    TripleM ($\mathbf{y}$,$\mathbf{y}^*$)   &0.815 &0.762 &0.069 &0.047 \\
    \midrule
    R-L ($\mathbf{y}$,$\mathbf{x}$)      &-0.465 &0.427 &-0.137 &0.224 \\
    TripleM ($\mathbf{y}$,$\mathbf{x}$)  &-0.556 &0.440 &-0.197 &0.140 \\
    QA [B-512]      &-0.305 &0.666 &-0.069 &0.113 \\
    QA [L-4096]     &-0.422 &0.629 &-0.100 &0.107 \\
    Entail [B-512]  &0.453 &0.212  &0.119 &0.024 \\
    Entail [L-4096] &-0.515 &0.345 &-0.109 &-0.061 \\
    % \midrule
    % \multicolumn{5}{c}{Supervised Without Reference Summary} \\
    \midrule
    CNN (weakly)     &0.685 &0.565 &0.198 &0.021 \\
    CNN              &0.838 &0.889 &0.315 &0.191 \\
    BERT             &0.907 &0.841 &0.267 &0.180 \\
    Longformer       &0.922 &0.926 &0.295 &0.203 \\
    \bottomrule
  \end{tabular}}
  \caption{Pearson's $r$ (complementary to Tab. \ref{tab:correlation_results}).}
  \label{tab:correlation_results_pearson}
\end{table}
% \section{Example Appendix}
% \label{sec:appendix}

% This is an appendix.

\end{document}